\UseRawInputEncoding
\pdfoutput=1 
\documentclass[unnumsec,webpdf,contemporary,large]{oup-authoring-template}%
\usepackage{caption}
\usepackage{graphicx}
\usepackage{makecell}
\usepackage{amsmath}
\usepackage{xspace}

\usepackage{rotating}
\usepackage{natbib}
\setcitestyle{numbers,square}

\theoremstyle{thmstyleone}

\theoremstyle{thmstyletwo}%
\theoremstyle{thmstylethree}%

\begin{document}

\journaltitle{Journal Title Here}
\DOI{DOI HERE}
\copyrightyear{2023}
\pubyear{2023}
\access{Advance Access Publication Date: Day Month Year}
\appnotes{Review}

\firstpage{1}


\title[Short Article Title]{Comprehensive evaluation of deep and graph learning on drug-drug interactions prediction}

\author[]{Xuan Lin}
\author[]{Lichang Dai}
\author[]{Yafang Zhou}
\author[]{Zu-Guo Yu}
\author[]{Wen Zhang}
\author[]{Jian-Yu Shi}
\author[]{Dong-Sheng Cao}
\author[]{Li Zeng}
\author[]{Haowen Chen}
\author[]{Bosheng Song}
\author[]{Philip S. Yu}
\author[]{Xiangxiang Zeng}

\authormark{Xuan et al.}

\corresp[]{Corresponding author: Haowen Chen and Bosheng Song, College of Computer Science and Electronic Engineering, Hunan University, 410013 Changsha, P. R. China. Email: hwchen@hnu.edu.cn, boshengsong@hnu.edu.cn}

\received{Date}{0}{Year}
\revised{Date}{0}{Year}
\accepted{Date}{0}{Year}

\abstract{Recent advances and achievements of artificial intelligence (AI) as well as deep and graph learning models have established their usefulness in biomedical applications, especially in drug-drug interactions (DDIs). DDIs refer to a change in the effect of one drug to the presence of another drug in the human body, which plays an essential role in drug discovery and clinical research. DDIs prediction through traditional clinical trials and experiments is an expensive and time-consuming process. To correctly apply the advanced AI and deep learning, the developer and user meet various challenges such as the availability and encoding of data resources, and the design of computational methods. This review summarizes chemical structure based, network based, NLP based and hybrid methods, providing an updated and accessible guide to the broad researchers and development community with different domain knowledge. We introduce widely-used molecular representation and describe the theoretical frameworks of graph neural network models for representing molecular structures. We present the advantages and disadvantages of deep and graph learning methods by performing comparative experiments. We discuss the potential technical challenges and highlight future directions of deep and graph learning models for accelerating DDIs prediction.}

\maketitle

\section{Introduction}
\textcolor{black}{Polypharmacy is progressively becoming the prevalent therapy by a patient for one or more conditions, especially for older patients with many chronic health conditions, and this trend continues to grow because of aging populations. For example, 67\% of elderly Americans were taking five or more medications \cite{kantor2016trends}. This can be baffling because potential drug interactions can alter the intended responses when patients taking multiple drugs simultaneously, which results in unexpected side effects or decreases clinical efficacy \cite{jin2017multitask}. These unintended interactions are widely referred to as drug-drug interactions (DDIs). As a common problem during polypharmacy, DDIs are associated with about 30\% of all reported adverse drug effects that becomes one of the most leading causes of trial failures in drug discovery and clinical research \cite{tatonetti2012novel, vilar2014similarity}. Take Ondansetron (Zofran) and dofetilide (Tikosyn) as an example. The former is a medication used to prevent nausea and vomiting, and the latter is used for heart rhythm. When they are used together, the amount of time between heartbeats can get too long. This can lead to dizziness, fainting, and even death in severe cases. As a result, predicting potential DDIs in advance is crucial for drug development and pharmacovigilance.}

\begin{figure*}[htbp]
\centering
\includegraphics[scale=1.0]{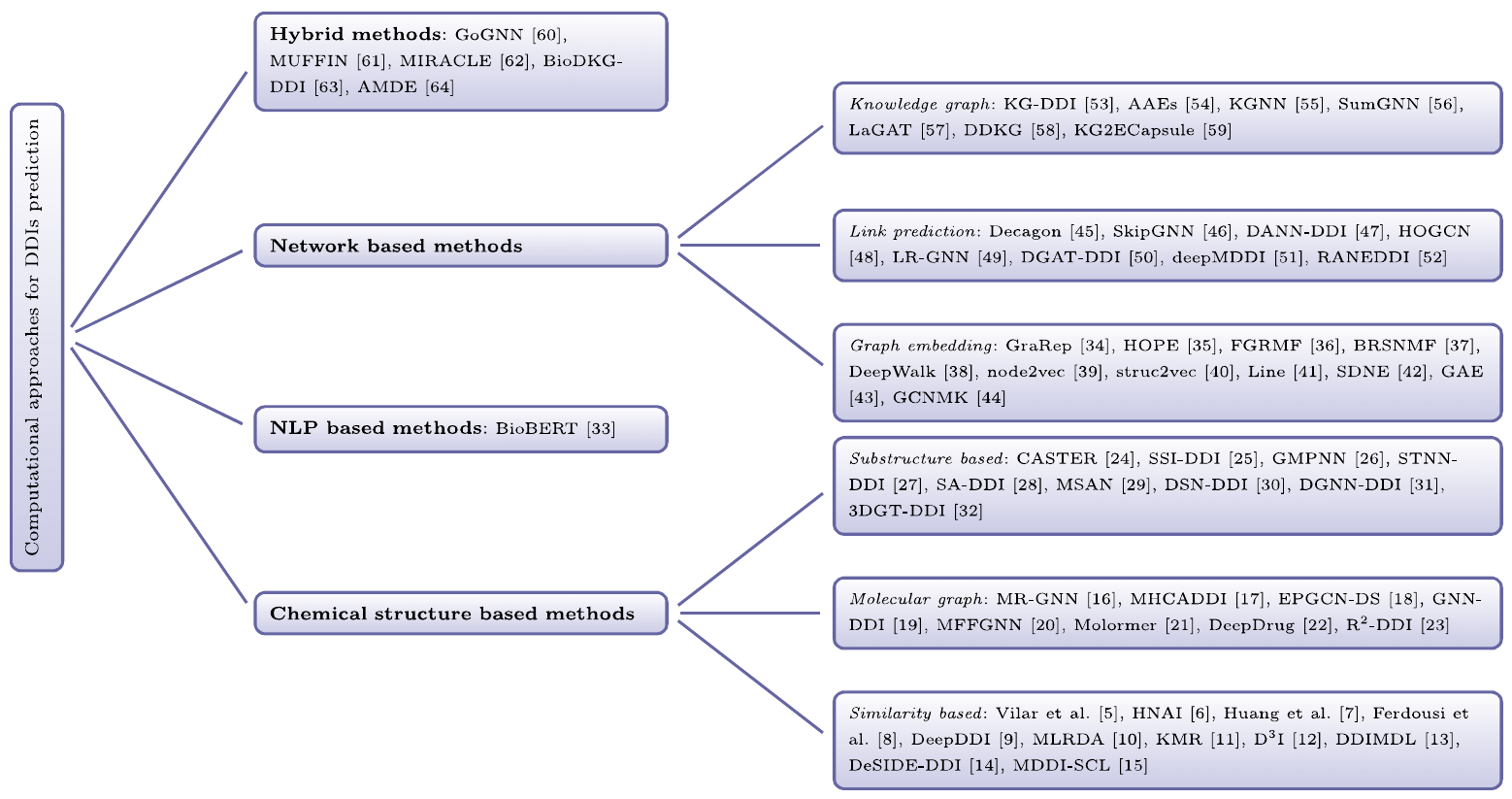}
\caption{The taxonomy of computational methods with representative examples. The computational approaches can be divided into chemical structure based, network based, NLP based and hybrid method, respectively. \textit{Chemical structure based methods} learn the structural and chemical features of molecules by using the similarity functions. These methods can be further classified into similarity based, molecular graph based and substructure based methods. \textit{Network based methods} that mainly rely on the topological characteristics of drugs in related networks integrate biological information from different data sources, which can be further classified into graph embedding, link prediction and knowledge graph based methods. \textit{NLP based methods} adopt the unsupervised and pretraining strategies in natural language processing (NLP) to learn contextual information from large unlabeled molecular datasets. \textit{Hybrid methods} combine with multiple types of features in an efficient pattern.}
\label{fig1}
\end{figure*}

\textcolor{black}{Identifying the existence of DDIs is the first step to avoid the potential adverse effects. Generally, DDIs can be broadly categorized into pharmaceutical, pharmacokinetic (PK) or pharmacodynamic (PD). Importantly, DDIs that primarily cause a change in PK will consequently lead to a secondary alteration in its PD. Thus, classifying DDIs' types is the further study that typically has been performed through extensive experimental testing in pharmaceutical research \cite{bansal2014community}, which can help the scientific communities and manufacturers further decrease toxicity and increase effectiveness for these interactions \cite{2016Computerized}. However, more direct costs are incurred during long-period of clinical trials. Besides, DDIs have important relation with combination therapies that are hard to explore the space of combinations via high-throughput screening due to the exceedingly large number of unique chemical combinations. Collectively, these considerations highlight the benefits of discovering novel computational methods for predicting DDIs.}

In the past few years, people have seen a surge in DDIs research due to the unprecedented success of deep and graph learning. Various databases, approaches and models have been proposed in the recent literature, urgently calling for a comprehensive survey to focus the efforts in this flourishing new direction. Few review articles cover machine learning algorithms and recently developed deep and graph learning models. Some surveys of databases and other resources supporting drug discovery and DDIs extraction have also been conducted recently \cite{qiu2021comprehensive, dong2022deep, pan2022deep, zeng2022deep, zeng2022toward, zhang2020deep}. In this survey, we first summarize the commonly adopted databases and molecular representations related to DDIs. Then we present an overview of computational methods for DDIs prediction and focus on reviewing deep learning and graph neural network (GNN) based methods. Moreover, we introduce several commonly-used GNN models. Furthermore, we select several representations baselines for comparative experiments on two benchmark datasets, and the detailed experimental results are analyzed. Finally, we make a conclusion to discuss the potential future trends as well as promising research directions that could be used to further improving DDIs prediction.
To summarize, the main contributions of this work are as follows:
\begin{itemize}
  \item \textit{Structured taxonomy}. As shown in Fig. 1, we contribute a structured taxonomy to provide a broad overview of computational methods, which categorizes existing works from four perspectives: chemical structure based, network based, NLP based and hybrid methods.
  \item \textit{Current progress}. We systematically delineate the current research directions on the topic of deep and graph learning methods for DDIs prediction as illustrated in Table 1, and we further investigate the comparison performance of these representative baseline models as shown in Tables 3-5.
  \item \textit{Abundant resources.} We have gathered a comprehensive collection of resources dedicated to DDIs prediction. These collections include open-sourced deep and graph learning methods, available platform and toolkit, as well as an important paper list. These resources can be accessed our github\footnote{\url{https://github.com/xzenglab/resources-for-DDIs-prediction-using-DL}}, which will be continuously updated.
  \item \textit{Future directions}. We discuss the limitations of existing works and suggest several promising future directions.
\end{itemize}

\begin{sidewaystable*}[htbp]
\caption{List of deep and graph learning models for DDI prediction.} \label{tab1}%
\centering
\scalebox{0.85}{
\begin{tabular}{ccccccc}
\toprule
\textbf{Model}   & \textbf{Year}  & \textbf{Input}  & \textbf{Representation} & \textbf{Architecture} & \textbf{Task} & \textbf{Code}\\
\midrule
DeepDDI \cite{ryu2018deep}  & 2018   & SMILES  & Structural similarity     & FC layers & Binary/multi-class classification &  \href{https://bitbucket.org/kaistsystemsbiology/deepddi}{\textcolor{blue}{Link}} \\
Decagon \cite{zitnik2018modeling} & 2018 & SMILES & Heterogeneous network & Encoder + decoder & Binary classification & \href{https://github.com/mims-harvard/decagon}{\textcolor{blue}{Link}} \\
MLRDA \cite{chu2019mlrda}   & 2019   & SMILES  & Drug features       & Encoder + decoder  & Binary classification &  -\\
KMR   \cite{shen2019kmr}    & 2019   & Drug ID & Multiple drug descriptor & CNN, Bi-LSTM + attention & Binary classification &  -\\
MR-GNN \cite{ijcai2019p551} & 2019   & SMILES  & Molecular graph     & Weighted GCN + LSTM & Binary/multi-class classification &
\href{https://github.com/prometheusXN/MR-GNN}{\textcolor{blue}{Link}}\\
MHCADDI \cite{deac2019drug} & 2019   & SMILES  & Molecular graph     & GCN + co-attention & Binary/multi-label classification &  \href{https://github.com/AstraZeneca/chemicalx}{\textcolor{blue}{Link}}\\
D$^3$I \cite{peng2019deep}  & 2019   & Drug ID & Drug features       & Encoder + aggregator  & Binary classification & \href{https://gitlab.com/peng10/d3i}{\textcolor{blue}{Link}} \\
KG-DDI \cite{karim2019drug} & 2019 & Drug ID & Knowledge graph & Conv-LSTM & Binary classification &
\href{https://github.com/rezacsedu/Drug-Drug-Interaction-Prediction}{\textcolor{blue}{Link}} \\
DDIMDL \cite{deng2020multimodal}& 2020 & Drug name & Diverse drug features & DNN & Multi-class classification &  \href{https://github.com/YifanDengWHU/DDIMDL}{\textcolor{blue}{Link}} \\
GoGNN \cite{ijcai2020p183}  & 2020   & SMILES  & Molecular/interaction graph     & GNN + attention & Multi-class/multi-label classification &  \href{https://github.com/Hanchen-Wang/GoGNN}{\textcolor{blue}{Link}} \\
KGNN \cite{lin2020kgnn}     & 2020   & Drug ID & Knowledge graph     & GNN & Binary classification &  \href{https://github.com/xzenglab/KGNN}{\textcolor{blue}{Link}} \\
BioBERT \cite{lee2020biobert} & 2020 & SMILES  & Embedding        & BERT   & Binary classification &
\href{https://github.com/dmis-lab/biobert}{\textcolor{blue}{Link}} \\
SkipGNN \cite{huang2020skipgnn} & 2020 & Drug ID & Skip graph & MPNN & Binary classification &  \href{https://github.com/kexinhuang12345/SkipGNN}{\textcolor{blue}{Link}} \\
CASTER \cite{huang2020caster} & 2020 & SMILES  & Substructure     & Encoder + decoder  & Binary classification &  \href{https://github.com/kexinhuang12345/CASTER}{\textcolor{blue}{Link}} \\
EPGCN-DS \cite{sun2020structure} & 2020 & SMILES & Molecular graph & GCN &  Binary classification &
\href{https://github.com/AstraZeneca/chemicalx}{\textcolor{blue}{Link}} \\
HOGCN \cite{kishan2021predicting} & 2021 & Drug ID & Interaction network & High-order GCN & Binary classification &  \href{https://github.com/kckishan/HOGCN-LP}{\textcolor{blue}{Link}} \\
MUFFIN \cite{chen2021muffin}  & 2021 & SMILES/Drug ID  & Molecular graph + knowledge graph    & MPNN + TransE & Binary/multi-class/multi-label classification &  \href{https://github.com/xzenglab/MUFFIN}{\textcolor{blue}{Link}} \\
SumGNN \cite{yu2021sumgnn}  & 2021 & SMILES/Drug ID  & Knowledge graph/subgraph   & GNN + attention & Multi-class/multi-label classification &  \href{https://github.com/yueyu1030/SumGNN}{\textcolor{blue}{Link}} \\
MIRACLE \cite{wang2021multi}  & 2021 & SMILES  & Molecular graph    & GCN + Contrastive learning & Binary classification &  \href{https://github.com/isjakewong/MIRACLE}{\textcolor{blue}{Link}} \\
SSI-DDI \cite{nyamabo2021ssi} & 2021 & SMILES  & Substructure     & GAT + Co-attention & Binary classification &  \href{https://github.com/kanz76/SSI-DDI}{\textcolor{blue}{Link}} \\
AAEs    \cite{dai2021drug}    & 2021 & Drug ID  & Knowledge graph  & Adversarial autoencoders & Binary classification &  \href{https://github.com/dyf0631/AAE_FOR_KG}{\textcolor{blue}{Link}} \\
GNN-DDI \cite{feng2022prediction}    & 2022     & SMILES & Molecular graph  & GAT & Binary classification & \href{https://github.com/NWPU-903PR/GNN-DDI}{\textcolor{blue}{Link}}\\
MFFGNN \cite{he2022multi}     & 2022 & SMILES + molecular graph   & Multi-type feature & GNN + BiGRU & Binary classification &  \href{https://github.com/kaola111/mff}{\textcolor{blue}{Link}} \\
GCNMK \cite{wang2022predicting} & 2022 & Drug ID  & DDI graph + drug features & GCN + Linear transformation  & Binary classification  & -\\
DeepDrug \cite{yin2022deepdrug} & 2022 & SMILES & Molecular graph & RGCN & Binary/multi-class/multi-label classification &
\href{https://github.com/wanwenzeng/deepdrug}{\textcolor{blue}{Link}} \\
LR-GNN \cite{kang2022lr}      & 2022 & Drug ID  & Biomedical network & GCN & Binary classification  & \href{https://github.com/zhanglabNKU/LR-GNN}{\textcolor{blue}{Link}}\\
DANN-DDI \cite{liu2022enhancing} & 2022 & Drug ID & Biomedical network & SDNE + attention & Binary classification  & \href{https://github.com/naodandandan/DANN-DDI}{\textcolor{blue}{Link}}\\
DGAT-DDI \cite{feng2022directed} & 2022 &  Directed graph &  Source/target encoding   &  Source/target GAT  &  Binary classification  & \href{https://github.com/F-windyy/DGATDDI}{\textcolor{blue}{Link}}\\
GMPNN \cite{nyamabo2022drug} & 2022 & SMILES & Molecular graph & Gated MPNN & Binary classification &
\href{https://github.com/kanz76/GMPNN-CS}{\textcolor{blue}{Link}} \\
STNN-DDI \cite{yu2022stnn}   & 2022 & SMILES & Substructure & Encoder + decoder  & Binary classification &
\href{https://github.com/zsy-9/STNN-DDI}{\textcolor{blue}{Link}} \\
deepMDDI \cite{feng2022deepmddi}& 2022 & Drug ID & Sub-networks & RGCN Encoder + decoder & Multi-label classification & \href{https://github.com/NWPU-903PR/MTDDI}{\textcolor{blue}{Link}}\\
RANEDDI \cite{yu2022raneddi}    & 2022 & Drug ID & DDI network  & RotatE + network embedding & Binary/multi-class classification &
\href{https://github.com/DongWenMin/RANEDDI}{\textcolor{blue}{Link}} \\
DeSIDE-DDI \cite{kim2022deside} & 2022 & Fingerprints &  Gene expressions & DNN & Multi-class classification & \href{https://github.com/GIST-CSBL/DeSIDE-DDI}{\textcolor{blue}{Link}} \\
SA-DDI \cite{yang2022learning}  & 2022 & SMILES  & Substructure  & D-MPNN & Binary classification  & \href{https://github.com/guaguabujianle/SA-DDI}{\textcolor{blue}{Link}}\\
MSAN \cite{zhu2022molecular}    & 2022 & SMILES  & Substructure  &  Transformer-like framework &  Binary classification  & \href{https://github.com/Hienyriux/MSAN}{\textcolor{blue}{Link}}\\
LaGAT \cite{hong2022lagat}      & 2022 & Drug ID   & Knowledge graph/subgraph  & Link-aware GAT  & Binary/multi-class classification &\href{https://github.com/Azra3lzz/LaGAT}{\textcolor{blue}{Link}} \\
Molormer \cite{zhang2022molormer} & 2022 & 2D structures & Molecular graph spatial structure & Attention + Siamese network & Binary classification & \href{https://github.com/IsXudongZhang/Molormer}{\textcolor{blue}{Link}} \\
MDDI-SCL \cite{lin2022mddi}    & 2022  &  Drug ID   & Drug features  & Attention + Contrastive learning  & Multi-class classification &\href{https://github.com/ShenggengLin/MDDI-SCL}{\textcolor{blue}{Link}} \\
R$^2$-DDI \cite{lin2022r2}     & 2022  & SMILES     & Molecular graph  & DeeperGCN + Feature refinement & Binary classification
&\href{https://github.com/linjc16/R2-DDI}{\textcolor{blue}{Link}} \\
BioDKG-DDI \cite{ren2022biodkg}& 2022  & SMILES & Multiple drug features  & Attention + DNN  & Binary classification & - \\
AMDE \cite{pang2022amde}       & 2022  & SMILES & Sequence + Atomic graph & MPAN + Transformer & Binary classification
&\href{https://github.com/wan-Ying-Z/AMDE-master}{\textcolor{blue}{Link}} \\
DDKG \cite{su2022attention}    & 2022  & SMILES/Drug ID & Knowledge graph & Encoder-decoder + GCN & Binary classification
&\href{https://github.com/Blair1213/DDKG}{\textcolor{blue}{Link}} \\
3DGT-DDI \cite{he20223dgt}     & 2022  & 3D structures & Molecular graph + position information & 3D GNN + text attention & Binary/multi-class classification & \href{https://github.com/hehh77/3DGT-DDI}{\textcolor{blue}{Link}} \\
DSN-DDI \cite{li2023dsn}       & 2023  & Molecular graph & Substructure & Dual-view encoder + decoder & Binary classification
& \href{https://github.com/microsoft/Drug-Interaction-Research/tree/DSN-DDI-for-DDI-Prediction}{\textcolor{blue}{Link}} \\
DGNN-DDI \cite{ma2023dual}     & 2023  & SMILES   & Molecular graph + substructure  & Directed MPNN + substructure attention     & Multi-class classification &\href{https://github.com/mamei1016/DGNN-DDI}{\textcolor{blue}{Link}} \\
KG2ECapsule \cite{su2022biomedical} & 2023 & Drug ID  & Knowledge graph   & GCN + Capsule  & Multi-label classification & \href{https://github.com/Blair1213/KG2ECapsule}{\textcolor{blue}{Link}} \\
\botrule
\end{tabular}}
\end{sidewaystable*}

\begin{table*}[!t]
\caption{The widely used databases for DDI prediction.} \label{tab2}%
\centering
\begin{tabular}{cccccc}
\toprule
\textbf{Database} & \textbf{Publication year} & \textbf{Num. of drug} & \textbf{Num. of drug-related pairs} & \textbf{Latest update} & \textbf{Link}\\
\midrule
KEGG \cite{kanehisa2022kegg}        & 1995 & 11,147 & 324,183 DDIs   & V104.1, 2022-11-01   & \href{https://www.genome.jp/kegg/drug/}{\textcolor{blue}{Link}}\\
DrugBank \cite{wishart2018drugbank} & 2006 & 1,706  & 191,808 DDIs   & V5.1.9, 2022-01-03   & \href{https://go.drugbank.com/}{\textcolor{blue}{Link}}\\
SIDER \cite{kuhn2016sider}          & 2008 & 1,430  & 139,756 drug-side effect pairs   & V4.1, 2015-10-21     & \href{http://sideeffects.embl.de}{\textcolor{blue}{Link}}                                   \\
TWOSIDES \cite{tatonetti2012data}   & 2012 &  645   & 4,649,441 DDIs & -                    & \href{https://tatonettilab.org/resources/nsides/}{\textcolor{blue}{Link}} \\
OFFSIDES \cite{tatonetti2012data}   & 2012 & 1,332  & 18,842 drug-event associations    & -                    & \href{https://tatonettilab.org/resources/nsides/}{\textcolor{blue}{Link}}        \\
BIOSNAP \cite{zitnik2018biosnap}    & 2018 & 1,332  & 41,520 DDIs    & -                    &
\href{http://snap.stanford.edu/biodata/}{\textcolor{blue}{Link}}        \\
\botrule
\end{tabular}
\end{table*}

\section{Data Sources}\label{Sec:data}
\color{black}{DDIs is not only known as a binary relationship but also can be affected by numerous factors, such as chemical substructures, targets and enzymes. The available datasets collect multiple drug-related information including mechanism of actions, protein structures and pharmacogenomic effects, which provide great opportunity for scientific communities to effectively develop novel methods to predict various drug interactions. In this section, we provide a brief overview of several commonly-used chemical and bioinformatics databases for DDIs prediction as shown in Table \ref{tab2}.}

\subsection{\textit{\textbf{KEGG}}}
\textcolor{black}{KEGG database is originally used to discover utilities of the biological system and high-level functions, especially large-scale molecular datasets generated by genome sequencing and other high-throughput experimental technologies. As an integrated database with sixteen resources, it was broadly classified into systems, genomic, chemical and health information, such as KEGG PATHWAY and KEGG DRUG. As for KEGG DRUG, it collects multiple drug information of approved drugs and unifies them according to their chemical structures. Specifically, each entry is identified by the drug number and associated with KEGG original annotations (e.g., drug metabolism), which results in 1,925 approved drugs and their 56,983 interactions spanning 11,147 drugs and 324,183 interactions respectively.}

\subsection{\textit{\textbf{DrugBank}}}
\textcolor{black}{DrugBank is a free-to-access and online database that collects drugs, drug targets, their mechanisms and interactions. Version 1.0 started in 2006 and the latest version has been updated to 5.1.9 in 2022. At present, it contains 14,944 drug entries, including 2,729 approved small molecule drugs, 1,564 approved biologics (e.g., proteins and allergenics) and over 6,713 experimental drugs including discovery-phase. Generally, given two drugs with their SMILES sequences, the final goal is to predict their interaction type (i.e., binary, multi-class and multi-label classification). DrugBank V5.1.4 is widely-used in comparison experiment, and it contains 1,706 drugs and 191,808 drug pairs with 86 DDI types.}

\subsection{\textit{\textbf{SIDER}}}
\textcolor{black}{Side Effect Resource collects multiple information from marketed drugs and their side effects to provide a more comprehensive view of actions of drugs and their adverse reactions. It can predict the potential side effects of drug candidates according to their binding fingerprints, chemical structures and other chemical properties. Meanwhile, it combines side effect information with other resources in chemical biology, which will greatly benefit pharmacology and medical research. Its current version 4.1 includes 1,430 drugs, 5,868 side effects and 139,756 drug-side effect pairs.}

\subsection{\textit{\textbf{TWOSIDES}}}
\textcolor{black}{The TWOSIDES databases collect polypharmacy side effects that are related to individual one in the drug pairs or higher-order drug combinations. Overall, it contains 868,221 associations between 59,220 pairs of drugs and 1,301 adverse events. Additionally, it contains 3,782,910 significant associations for which the drug pair has a higher side-effect association score, evaluated by the proportional reporting ratio (PRR) \cite{rothman2004reporting}, than those of the individual drugs alone. Specifically, it contains 645 drugs and side effects caused by 63,473 combinations of different drugs. Generally, given two drugs with their SMILES sequences, the final goal is to predict all side effects (i.e., multi-label classification).}

\subsection{\textit{\textbf{OFFSIDES}}}
\textcolor{black}{The OFFSIDES database collects 438,801 off-label side effects between 1,332 drugs and 10,097 adverse events. Off-label means no record on the US Food and Drug Administration (FDA)'s official drug label while on-label means the opposite. The drug label lists in average 69 on-label adverse events. And it listed an average of 329 high-confidence off-label adverse events for each drug. Moreover, it recovers 38.8\% (i.e., 18,842 drug-event associations) of SIDER associations from the adverse event reports.}

\subsection{\textit{\textbf{BIOSNAP}}}
\textcolor{black}{The BIOSNAP dataset collects various types of interactions between FDA-approved drugs by constructing a biological network. Nodes represent drugs and edges represent drug interactions. This dataset contains 1,322 approved drugs with 41,520 labelled DDIs that are extracted from drug labels and scientific publications.}

\section{Molecular Representation}\label{Sec:molecular}
\begin{figure*}[htbp]
\centering
\includegraphics[scale=0.3]{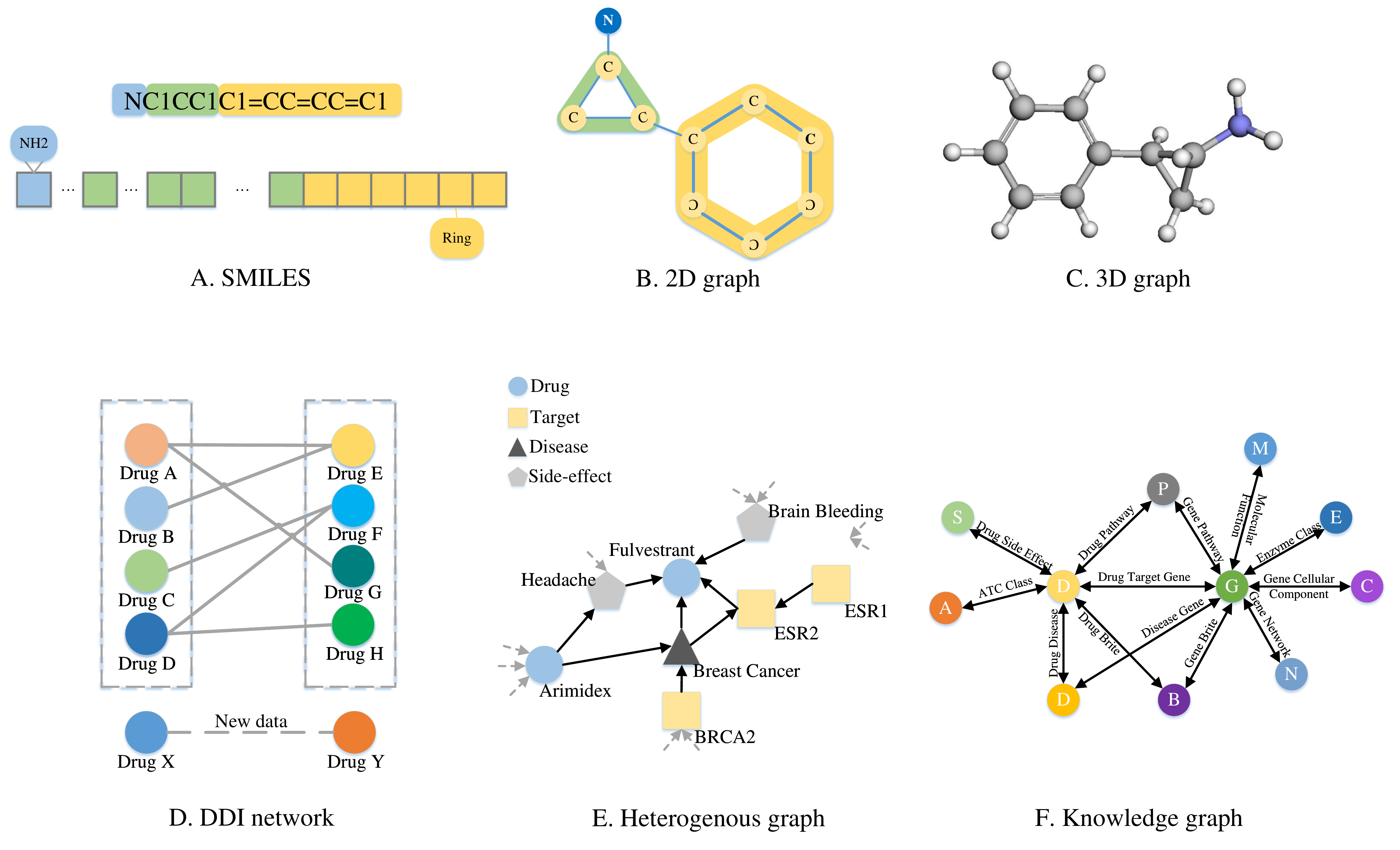}
\caption{A diagram illustrating six commonly-used molecular representation approaches, including: (A) one-dimensional (1D) sequence-based representation; (B) 2D graph-based representation; (C) 3D representation; (D) DDI network; (E) heterogeneous graph and (F) knowledge graph.}
\label{fig2}
\end{figure*}

\textcolor{black}{The representation of drug molecule is an crucial part in drug-related tasks, including DDIs prediction. For example, Tranylcypromine is an inhibitor of the enzyme monoamine oxidase \cite{baldessarini2006drug}, functioning nonselectively and irreversibly, and thus it is also employed clinically as an antidepressant and anxiolytic agent in the treatment of mood and anxiety disorders. Take its SMILES (Simplified Molecular Input Line Entry System) \cite{weininger1988smiles} format \textit{C1C(C1N)C2=CC=CC=C2} as an example, Tranylcypromine is represented by six commonly-used molecular representation as shown in Fig. \ref{fig2}.}

\subsection{\textit{\textbf{Sequences}}}
\textcolor{black}{As the most frequently-used molecule descriptor, SMILES is a string of characters as shown in Fig. \ref{fig2}(A), where each atom is encoded by a respective ASCII symbol, and chemical bonds, branching as well as stereochemistry are represented by specific symbols in SMILES strings. The SMILES sequence is capable of converting the chemical structure into a spanning tree by utilizing a longitudinal-first traversal tree algorithm to generate a sequence of characters. A variety of deep learning models, such as recurrent neural networks, are able to employ their internal state (memory) to process variable length sequences of inputs \cite{sutskever2014sequence, xu2017seq2seq, zhang2018seq3seq}, using SMILES sequences as input to extract the chemical context via various natural language processing techniques, including Mol2Vec \cite{jaeger2018mol2vec} and FCS \cite{huang2021moltrans}. Sequence-based representations tend to be compact, memory-efficient, and easily searchable.}

\subsection{\textit{\textbf{2D graph}}}
\begin{figure*}[htbp]
\centering
\includegraphics[scale=0.2]{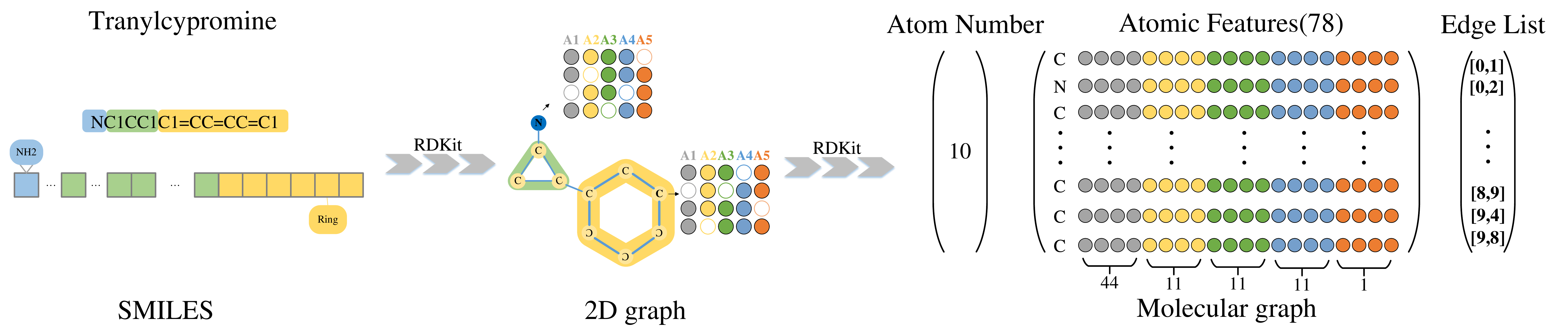}
\caption{An example of the molecular graph representation of \textit{Tranylcypromine} by using RDKit.}
\label{fig3}
\end{figure*}

A more direct way to representing drug molecules is through 2D graph-based representation (i.e., molecular graph) as shown in Fig. 2(B). In particular, we denote 2D graph as $G_{2D} = (X, E)$, $X \in \mathbb{R}^{N \times d}$ represents the atom attribute matrix, where $N$ denotes the number of nodes and $d$ denotes the dimensionality of node feature, and $E$ are characterized by the type of chemical bonds between the atoms, including single, double, triple and aromatic bond. Specifically, Figure \ref{fig3} shows an example of the molecular graph representation of \textit{Tranylcypromine}. First, the SMILES sequence is transformed into its 2D structures using RDKit tool. Predefined atomic features are then assigned to each node based on its atom number. In a molecular graph, each node contains a 78-dimension initial feature vector to encode 5 types of atomic features, including atomic symbol, adjacent atoms, adjacent hydrogens, implicit value and aromaticity. Finally, we obtain the molecular graph representation of \textit{Tranylcypromine} that consists of atom number (i.e., total number of atoms), atomic features and edge features (i.e., edge list). This representation allows us to extract the structural information from a molecular graph. We then typically apply a transformation function $T_{2D}$ to the topological graph. Given a 2D graph $G_{2D}$ or molecular graph obtained from its SMILES sequence via RDKit \cite{landrum2006rdkit}, its representation $H_{2D}$ can be computed from a 2D GNN model:
\begin{equation}
H_{2D} = 2D GNN(T_{2D}(X, E)).
\end{equation}
\textcolor{black}{Usually, message passing neural networks (MPNN) \cite{gilmer2017neural} known as one of the classic 2D GNN models are designed to accomplish the encoding of graph-based methods. Since 2D graph is usually stored in the form of adjacency matrices. The utilization of 2D GNN not only allows for faster and accurate combination of properties between two adjacent atoms or chemical bonds, but it also allows the weights to be optimized in the message passing process. With comparison to sequence-based approaches, graph-based representations are easy to extract the structural information via graph convolutional operations, where bond weights can be updated and optimized in message-passing networks.}

\subsection{\textit{\textbf{3D graph}}}
\textcolor{black}{Many methods use sequence- and graph-based representations of molecules as the inputs, such as SMILES sequences and molecular graphs. Although these methods can effectively preserve the structural information of drug molecules, they can not capture well the inter-binding relationships between ligands and receptors, especially biologically meaningful in 3D relationships, because the 3D coordinates of all atoms in the ground-state molecule are critical to various applications, including molecular property prediction \cite{li2022geomgcl} and molecular conformer ensembles \cite{ganea2021geomol}. Here, 3D graph as shown in Fig. \ref{fig2}(C) represents the spatial arrangements of each atom in the 3D space, containing a list of atoms with atom types and atomic coordinates. Generally, each molecule with $n$ atoms is expressed as an undirected graph $\mathcal{G} = (\mathcal{V}, \mathcal{E})$, where $\mathcal{V}=\{v_i\}^n_{i=1}$ is the set of vertices symbolizing atoms and $\mathcal{E} = \{e_{ij} \vert (i, j) \in \lvert\mathcal{V}\rvert \times \lvert\mathcal{V}\rvert\}$ is the set of edges representing inter-atomic bonds. Each node $v_i \in \mathcal{V}$ indicates the atomic attributes (e.g., the element type), and each edge $e_{ij} \in \mathcal{E}$ describes the connection between $v_i$ and $v_j$, and is labeled with its chemical type. Additionally, we also assign virtual types to the unconnected edges. For 3D geometry graph, each atom in $\mathcal{V}$ is embedded in the 3-dimensional space with a coordinate vector $\textbf{c} \in \mathbb{R}^3$, and the full set of positions (i.e., the conformation where atoms are represented as their Cartesian coordinates) can be represented as $C = [\textbf{c}_1, \textbf{c}_2, ..., \textbf{c}_n]$, where $\textbf{c}_i \in \mathbb{R}^3$. Then we generally apply a transformation function $T_{3D}$ on the geometry graph. Given a geometry graph $G_{3D} = (X, C)$, its representation $H_{3D}$ can be obtained via a 3D GNN model:}
\begin{equation}
H_{3D} = 3D GNN(T_{3D}(X, C)).
\end{equation}

\subsection{\textit{\textbf{DDI network}}}
\textcolor{black}{DDIs can be associated with biological, chemical, and phenotypic information about drugs. The drug-drug interaction network (DDI network) is proposed to learn the potential associations between drug molecules. Generally, the DDIs prediction problem is formulated as a missing link prediction task by constructing a DDI network with drugs as nodes and known interactions as edges. Specifically, drugs should be represented as feature vectors via interaction profile from known interactions to build prediction models. As shown in Fig. \ref{fig2}(D), let \textit{A, B, C, ..., H} be a set of given drugs, the drug interaction profile, which is a binary vector indicating the presence or absence of interaction between drugs (e.g., \textit{A}$\longleftrightarrow$\textit{E}, \textit{A}$\longleftrightarrow$\textit{G}), can be represented as an interaction network.}

\subsection{\textit{\textbf{Heterogeneous graph}}}
\textcolor{black}{Heterogeneous graph (HetG) contains a wealth of information with structural relations (i.e. edges) among nodes of various types,
as well as unstructured content associated with each node \cite{zhang2019heterogeneous}. For example, HetG can be expressed as to involve many other types of biological entity relationships in the process of predicting DDIs. Considering these different associations can enhance the prediction performance. In general, the HetG associated with DDIs is expressed as a graph $G = (V, E, O_V, R_E)$, where $V$ and $E$ denote the sets of nodes and links, respectively. $O_V$ and $R_E$ represent the set of object types and that of relation types, respectively. Furthermore, each node is associated with heterogeneous contents (e.g., attributes). Specifically, the HetG denotes relations between different pairs, including drugs and targets, drugs and side-effects, drugs and diseases. For instance, Fig. \ref{fig2}(E) illustrates the biological heterogeneous graph centered around drug \textit{Fulvestrant}, where edges with distinct colors denote different relations, and arrows indicate the direction of information flow.}

\begin{figure*}[htbp]
\centering
\includegraphics[scale=0.25]{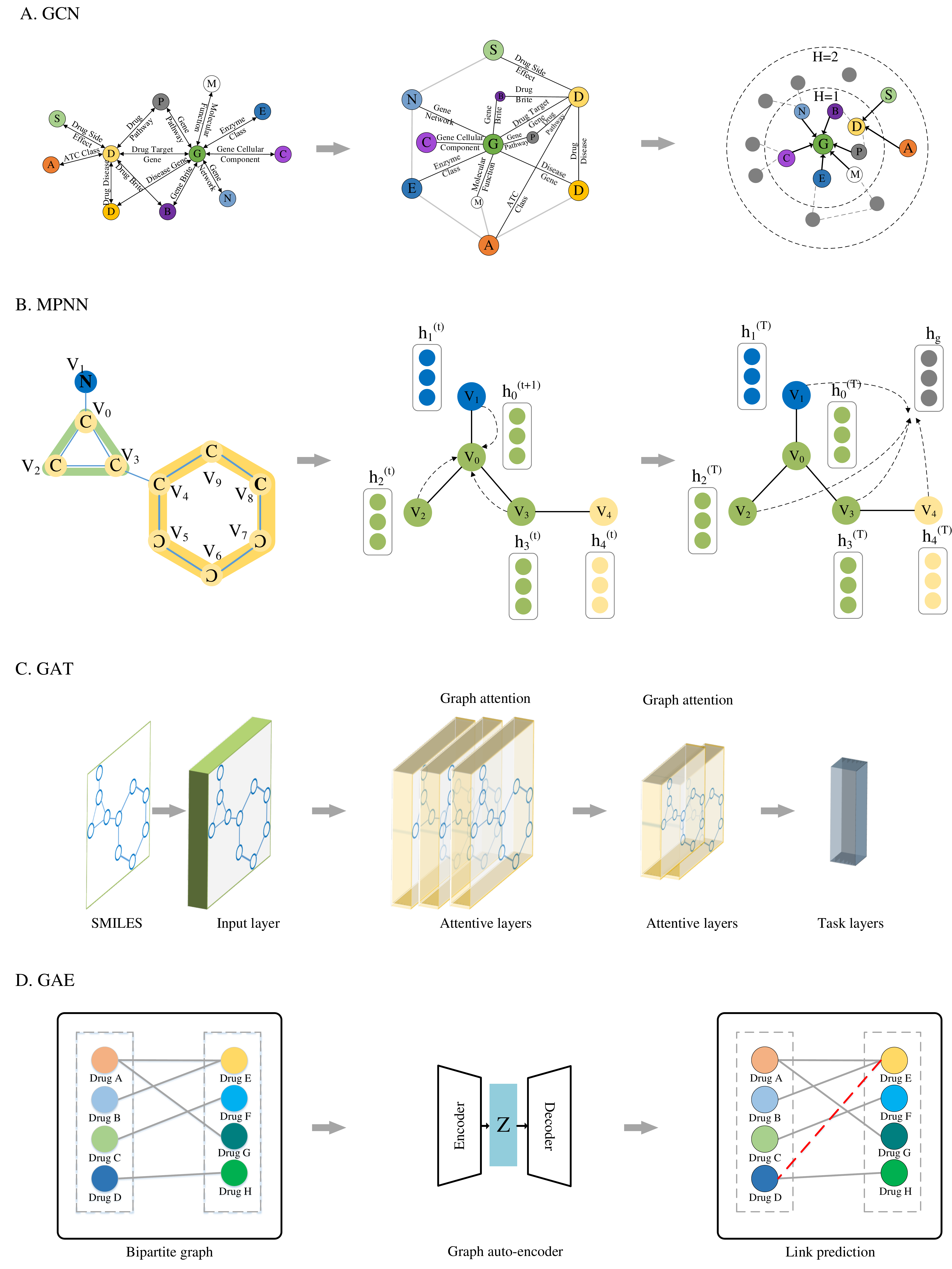}
\caption{A diagram illustrating the theory framework of widely-used GNN models in the DDIs prediction, including: (A) GCN; (B) MPNN; (C) GAT; (D) GAE.}
\label{fig4}
\end{figure*}

\subsection{\textit{\textbf{Knowledge graph}}}
\textcolor{black}{Recently, knowledge graphs (KGs), which are a form of structured human knowledge, have been gaining increasing attention from both academic and various aspects of the drug discovery domain \cite{ji2021survey}. For example, KGs can be utilized to better integrate multiple entity types and diverse association relations between biological entities. This approach allows for the extraction of high-order semantic features to improve DDIs prediction. KGs can be represented by a structured representation of facts, which comprised of entities, relationships, and semantic descriptions. Generally, a knowledge graph is denoted by $\mathcal{G} = (\mathcal{V}, \mathcal{E}, \mathcal{F})$, where $\mathcal{E}$, $\mathcal{R}$ and $\mathcal{F}$ are sets of entities, relations and facts, respectively. A fact is denoted as a triple $(h, r, t) \in \mathcal{F}$. As shown in Fig. \ref{fig2}(F), entities (i.e., nodes) with different color and alphabet represent real-world biological objects (e.g., drug, target and side-effect), and relationships (i.e., edges) depict the connection between entities, where semantic descriptions of entities and their relationships encompass types and properties with a clearly defined meaning, including \textit{Drug Disease}, \textit{Drug Target Gene}, and \textit{Drug Brite}. As a topical concrete application, KGs have been utilized in helping to combat the COVID-19 pandemic \cite{domingo2021covid, reese2021kg}. Additionally, there are few existing knowledge graph covering various aspects of the drug discovery process, including Hetionet \cite{himmelstein2017systematic}, DRKG \cite{drkg2020}, BioKG \cite{walsh2020biokg}, PharmKG \cite{zheng2021pharmkg}, OpenBioLink \cite{breit2020openbiolink} and Clinical Knowledge Graph \cite{santos2020clinical}. Note that a comprehensive review is beyond the scope of this work and readers who are interested are directed to a dedicated review \cite{bonner2022review}.}

\section{Models}\label{Sec:models}
The DDIs prediction is to develop a computational model that receives two drugs with an interaction type as inputs and generates an output prediction indicating whether there exists an interaction between them. As introduced in previous section (i.e., Molecular Representation), drug molecules generally use RDKit to convert its SMILES sequence into molecular graphs with nodes as atoms and edges as chemical bonds. Specifically, a graph for a given SMILES is denoted by $G = (V, E)$, where $V$ is the set of $N$ nodes represented by a $d$-dimensional vector, and $E$ is the set of edges represented as an adjacency matrix $A$. In a molecule graph, $x_i$ (resp., $x_j$) $\in V$ is the $i$ (resp., $j$)-th atom and $e_{ij} \in E$ is the chemical bond between the $x_i$ and $x_j$. Owing to the non-Euclidean and translation invariance, GNNs have been proposed to replace traditional convolution networks in order to extract drug feature representations from chemical molecular graph. In the case of GNN, the process of learning drug representation is essentially the message passing between each node and its neighboring nodes. Thus we further systematically review four types of GNN models for encoding molecular representations into continuous vectors, as shown in Fig. \ref{fig3}, including graph convolutional network, message passing neural network, graph attention network and graph auto-encoder.

\subsection{\textit{\textbf{Graph convolutional networks}}}
\textcolor{black}{Graph convolutional networks (GCNs) are a class of neural networks specifically designed for graph-structured datasets. Advances in this direction are often categorized as spectral- and spatial-based approaches. Spectral-based approaches learn compact representations of graph elements, their attributes and supervised labels as shown in Fig. \ref{fig4}(A). Following the original paper of GCN \cite{kipf2016semi}, the input of multi-layer GCN is the node feature matrix $X \in \mathbb{R}^{N \times d}$ and the adjacency matrix $A \in \mathbb{R}^{N \times N}$ that represents the connection of nodes. The layer-wise propagation algorithm can be obtained as below:}
\begin{equation}
H^{l+1} = \sigma (\tilde{D}^{-1/2}\tilde{A}\tilde{D}^{-1/2}H^{(l)}W^{(l)}),
\end{equation}
\textcolor{black}{where $\tilde{A} = A + I_N$ is the adjacency matrix of an undirected graph $\mathcal{G}$ with added self-connections, and $I_N$ represents the identity matrix, $\tilde{D}$ is diagonal matrix with $\tilde{D}_{ii}=\sum\limits{_j}\tilde{A}_{ij}$ and $W^{(l)}$ is a layer-specific trainable weight matrix. Here, $\sigma(\cdot)$ denotes an activation function (e.g., $ReLU(\cdot) = max(0, \cdot)$). And $H^{(l)}$ (resp., $H^{(l+1)}$) $\in \mathbb{R}^{N \times D}$ represents the matrix of activations in the $l$ (resp., $l+1$)-th layer, respectively. We suppose that $H^{(0)} = X$. The output $Z \in \mathbb{R}^{N \times F}$ ($F$ is the number of output features for every node) can be obtained as below:}
\begin{equation}
Z = \sigma (\tilde{D}^{-1/2}\tilde{A}\tilde{D}^{-1/2}X\Theta),
\end{equation}
\textcolor{black}{where $\Theta \in \mathbb{R}^{F \times d}$ represents the matrix of filter parameters.}

In contrast, spatial-based approaches \cite{hamilton2017inductive} define convolutions directly on the graph by propagating and aggregating node representations from neighboring nodes in the vertex domain, as opposed to spectral-based GCN which depends on the specific eigenfunctions of the Laplacian matrix. Following KGNN \cite{lin2020kgnn}, the proposed method learns latent representations of drugs and their neighborhood entities embedding between drug pairs from the constructed KG. Fig. \ref{fig4}(A) shows an example of a 2-layer KGNN of the given \textit{G} node (green) in a KG. Note that besides the immediate neighbors (e.g., \textit{B}, \textit{E} and \textit{P}), it also extends KGNN to 2-layer ($H$=2)
to extract both high-order structures and semantic relations. Generally, given a node $v$ at the $k$-th depth and its graph convolution is computed by:
\begin{equation}
\textbf{h}^k_{\mathcal{N}(v)} \leftarrow \mathrm{AGGREGATE}_k \left( \{\textbf{h}^{k-1}_u, \forall u \in \mathcal{N}(v)\} \right),
\end{equation}
\begin{equation}
\textbf{h}^k_v \leftarrow \sigma \left(W^k \cdot \mathrm{CONCAT}(\textbf{h}^{k-1}_v, \textbf{h}^k_{\mathcal{N}(u)}) \right),
\end{equation}
\textcolor{black}{where each node $v \in V$ aggregates the representation vectors of all its immediate neighboring nodes $u \in \mathcal{N}(v)$ in the current depth via some learnable $\mathrm{AGGREGATE}$ operation. Then it combines the node's current representation $\textbf{h}^{k-1}_v$ with its aggregated neighborhood representation $\textbf{h}^{k-1}_{\mathcal{N}(v)}$, and finally passes the combined vector to a fully-connected layer with a nonlinear activation function $\sigma(\cdot)$, followed by a normalization step. And the output of final representation at depth $K$ are denoted by $\textbf{z}_v=\textbf{h}^K_v$.}

\textcolor{black}{The aggregator functions include $mean$, $LSTM$, and $pooling$ aggregators. The $mean$ aggregator can be simplified as follows:}
\begin{equation}
\textbf{h}^k_v \leftarrow \sigma \left(W \cdot \mathrm{MEAN}(\{\textbf{h}^{k-1}_v\} \cup \{\textbf{h}^{k-1}_u, \forall u \in \mathcal{N}(v)\}) \right),
\end{equation}
\begin{equation}
\mathrm{AGGREGATE}^{pooling}_k = \mathrm{max}\left(\{\sigma(W_{pool}\textbf{h}^k_{u_i}+b), \forall u_i \in \mathcal{N}(v)\}\right).
\end{equation}

\subsection{\textit{\textbf{Message passing neural network}}}
Message passing neural network (MPNN) is a typical type of GNNs that maps an undirected graph $\mathcal{G}$ to a graph-level vector $h_\mathcal{G}$ using \textit{Message passing} and \textit{readout}. As depicted in Fig. \ref{fig4}(B), \textit{message passing} is first used to update node-level features (i.e., $V_0$) by aggregating messages from their neighbor nodes (i.e., $V_1$, $V_2$ and $V_3$). Following that, the \textit{readout} process is designed to generate a graph-level feature vector by aggregating all the node-level features from a molecule graph. Finally, a label is predicted for the graph based on the graph-level feature vector. Concretely, the \textit{Message passing} consists of $T$ steps. On each step $t$, node-level hidden feature $h_i^{(t)}$ and messages $m_i^{(t)}$ associated with each node $v_i$ are updated using message function $M_t$ and node update function $U_t$. Their definitions are as follows.
\begin{equation}
m_i^{t+1} = \sum_{v_j \in N(v_i)}M_t(\textbf{h}_i^{(t)}, \textbf{h}_j^{(t)}, e_{ij}),
\end{equation}
\begin{equation}
\textbf{h}_i^{t+1} = U_t(\textbf{h}_i^{(t)}, m_i^{t+1}),
\end{equation}
where $N(v_i)$ represents the set of neighbors of $v_i$ in the graph $\mathcal{G}$, and $h_i^{(0)}$ is set to the initial atom features $x_i$. The \textit{readout} then uses a readout function $\mathrm{R}$ to obtain a graph-level feature vector based on the node-level features at the final step as follows.
\begin{equation}
\textbf{h}_\mathrm{g} =\mathrm{R}(\{\textbf{h}_i^{(T)} | v_i \in \mathcal{G}\}).
\end{equation}
The message function $M_t$, node update function $U_t$, and readout function $\mathrm{R}$ are all learned differentiable functions.

\subsection{\textit{\textbf{Graph attention network}}}
\textcolor{black}{Traditionally, GCN assigns the same weight to each neighbor node, and not every neighbor node has the same importance. Thus, graph attention network (GAT) is proposed to introduce a graph convolution model based on self-attention mechanism, which incorporates a \textit{graph attention layer} in its architecture as shown in Fig. \ref{fig4}(C). According to the original paper of GAT \cite{velivckovic2017graph}, a set of node features $x \in R^F$ is used as input of GAT layer, and a linear transformation is applied to each node based on a weight matrix $W \in R^{F \times F^{'}}$, where $F$ and $F^{'}$ are the dimensions of the input and output nodes, respectively. Moreover, \textit{attention coefficients} between a node and its 1-hop neighbors are adopted to obtain the output node as follows:}
\begin{equation}
    e_{ij} = \alpha(W\Vec{h}_i, W\Vec{h}_j),
\end{equation}
\textcolor{black}{where $e_{ij}$ represents the importance of node $j$ to node $i$. To ensure that the coefficients are comparable across different nodes, they are normalized across all choices of $j$ using the softmax function as follows:}
\begin{equation}
    \alpha_{ij} = softmax_j(e_{ij}),
\end{equation}
\textcolor{black}{the non-linearity function $\sigma$ is finally applied to compute the output node $\Vec{h}_i^{'}$ by:}
\begin{equation}
    \Vec{\textbf{h}}_i^{'} = \sigma(\sum_{j \in N_i} a_{ij}W\Vec{h}_j),
\end{equation}
\textcolor{black}{while a basic operation of $attention$ is multi-head. Simply, it is to repeat the previous operation multiple times, but the parameters that need to be trained are different each time, so that we can extract more information. The process of multi-head can be computed by:}
\begin{equation}
    \Vec{\textbf{h}}_i^{'} = \prod_{k=1}^K \sigma(\sum_{j \in N_i} a^k_{ij}W^k\Vec{\textbf{h}}_j),
\end{equation}
\textcolor{black}{where $K$} is the number of heads.

\subsection{\textit{\textbf{Graph auto-encoder}}}
Graph auto-encoder (GAE) has been widely used in the field of unsupervised learning on graph-structure data. Obtaining the suitable embeddings to represent nodes in the graph is not trivial, GAE adopt the encoder-decoder structure to realize the goal and to apply to the downstream tasks, such as link prediction. If we view drugs as nodes and DDI as links in a graph, DDIs prediction can be considered as a task to complete a DDI adjacency matrix. As shown in Fig. \ref{fig4}(D), the encoder represents drugs into scalars and decoders use these scalars to rebuild the whole graph by predicting the existence of a link between a pair of nodes/drugs. The encoder can be viewed as representation methods and decoder can be viewed as classifiers. Generally, GAE employs GCN as an encoder to obtain latent representations or embedding of nodes. This process can be expressed as follows:
\begin{equation}
    Z = \mathrm{GCN}(X, A),
\end{equation}
\textcolor{black}{where $Z$ represents the latent representations of all nodes, $X$ and $A$ represent the feature matrix of the node and adjacency matrix, respectively. Here $X$ and $A$ as input are then fed into $GCN$ function, and we have:}
\begin{equation}
    \mathrm{GCN}(X, A)=\tilde{A}ReLU(\tilde{A}XW_0)W_1,
\end{equation}
\textcolor{black}{where $\tilde{A}=D^{-1/2}AD^{-1/2}$, $W_0$ and $W_1$ represent parameters to be learned. In short, $GCN$ is equivalent to a function that takes node features and adjacency matrix as input and outputs node embedding. After that, GAE uses the inner-product as a decoder to reconstruct the original graph, the computation of reconstructed adjacency matrix $\hat{A}$ can be formulated by:}
\begin{equation}
    \hat{A}=\sigma(ZZ^T),
\end{equation}
\textcolor{black}{in order for the reconstructed adjacency matrix to be as close to the original adjacency matrix as possible. Because the adjacency matrix determines the structure of the graph.}

\begin{figure*}[htbp]
\centering
\includegraphics[scale=0.45]{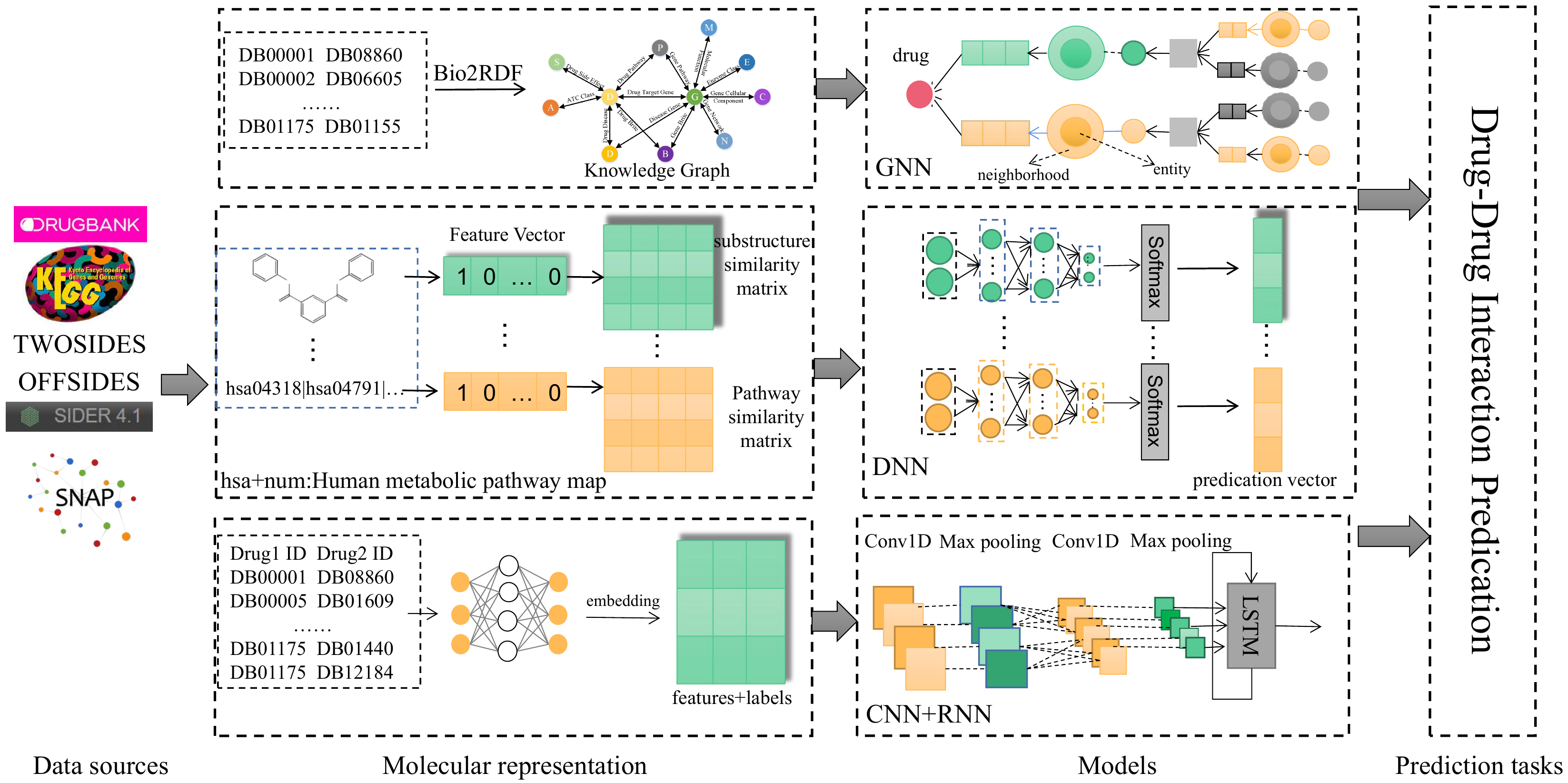}
\caption{The pipeline of deep and graph learning methods for DDIs prediction. In general, drugs and their relevant information such as ID, name, and SMILES sequences are obtained from accessible \textit{data sources} like DrugBank and TWOSIDES dataset. These drugs can then be optionally encoded into feature vectors using various \textit{molecular representation} methods, such as molecular graph representation. The resulting representations, such as similarity matrices or 2D graphs, are subsequently fed into suitable \textit{models}, such as Graph Neural Networks (GNNs), to generate interaction results or predicted scores based on the specific \textit{prediction tasks}.}
\label{fig5}
\end{figure*}

\section{Prediction tasks}
Due to the increasing amount of data and advanced algorithms, DL has led to breakthroughs in various domains \cite{gainza2020deciphering, mendez2020novo, beker2020minimal, jumper2021highly}, including in the application of DDIs and drug-related prediction tasks \cite{zeng2019deepdr, zeng2020target, yang2022modality, zeng2022accurate, ma2022kg}. Fig. \ref{fig5} shows an illustrative pipeline of several DL methods.

\textcolor{black}{In the beginning, this line of work develop effective representation method (see \textbf{Section \ref{Sec:molecular}}) to capture high-level hidden embeddings from various public datasets (see \textbf{Section \ref{Sec:data}}). Different from traditional machine learning based methods that heavily rely on the handcraft feature and domain knowledge, these approaches can learn more abstract information via deep architectures (see \textbf{Section \ref{Sec:models}}) without manually selecting and tuning features \cite{lee2019novel, lin2022mdf}, and the learned latent embeddings are finally used to predict on downstream tasks. There are many different types of classification tasks that may be encountered in DDIs prediction and specialized approaches to modeling that may be used for each, including binary, multi-class and multi-label classification.}

\textcolor{black}{Classification based predictive modeling involves assigning a class label to input sample. \textit{Binary classification} refers to predicting whether interactions exist without determining their specific type, and \textit{multi-class classification} involves predicting the specific type of DDIs between drug pairs. Following \cite{chen2021muffin} in the model training, we generally optimized the model parameters by minimizing the cross-entropy loss in the binary and multi-label classification tasks, as described below:}
\begin{equation}
    \mathcal{L}_1 = -[y_{ij}log\hat{y}_{ij} + (1-y_{ij})log(1-\hat{y}_{ij})],
\end{equation}
\textcolor{black}{where $\hat{y}_{ij}$ denotes the interaction label for drug pair ($d_i, d_j$) in binary classification task, and in multi-label task, each element $y_{ij}$ is the one-hot vector with 86 elements (e.g., 86 DDI types in DrugBank dataset).}

\textit{Multi-label classification} involves predicting one or more DDIs type for each drug pair, the loss is defined as follows:
\begin{equation}
    \mathcal{L}_2 = -\sum_{c=1}^{N_c}y_clog\hat{y}_c,
\end{equation}
\textcolor{black}{where $N_c$ is the number of multi-class DDI types, $y_c \in {0,1}$ describes whether current type $c$ is the same as the true label of sample pair, and $\hat{y}_c$ indicates the probability that the observed sample ($d_i, d_j$) belongs to type $c$.}

\section{Progress and Taxonomy of Computational Approaches}
Computational approaches mainly design effective algorithms to discover patterns by using public datasets retrieved from clinical texts \cite{kim2020ensemble}, electronic health records \cite{datta2021machine, wu2021ddiwas}, and social media \cite{vilar2018detection}. These methods can be roughly divided into chemical structure, network based, NLP based and hybrid methods.
A Taxonomy of the different methods is shown in Fig. \ref{fig1}. Furthermore, we summarize and formulate 46 state-of-the-art deep and graph learning models in the recent years using a unified symbolic system in Table \ref{tab1}.

\subsection{\textit{\textbf{Chemical structure based methods}}}
The vast majority of chemical structure based methods rely on similarity based, molecular graph, and substructure based approaches, respectively.
\subsubsection{Similarity based}
\textcolor{black}{Theses methods are based on the assumption that similar drugs may perform similar DDIs. They first extract some similarities from molecular structures \cite{vilar2012drug}, and various properties (e.g., phenotypic \cite{cheng2014machine}, functionality \cite{ferdousi2017computational}, and side effects \cite{huang2014systematic}) as features for model training. Then they adopt classifiers to predict potential DDIs. For example, DeepDDI \cite{ryu2018deep}, which consists of structural similarity profile generation pipeline and deep neural network (DNN), is proposed to use the structural information to classify 86 DDIs types. MLRDA \cite{chu2019mlrda} is proposed to effectively exploit multiple drug features by leveraging a novel unsupervised disentangling loss CuXCov. Similarly, a knowledge-oriented DNN model is developed by KMR \cite{shen2019kmr} to discover the interaction information among multiple features. Furthermore, D$^3$I \cite{peng2019deep} is presented to conduct cardinality- and order-invariant high-order DDIs prediction. DDIMDL \cite{deng2020multimodal} is constructed by the similarity assumption, and it built a multi-modal DL framework with multiple drug features to predict DDI events. A novel DL-based framework named DeSIDE-DDI \cite{kim2022deside} is developed to show more concern in interpretation on underlying genes, and it leveraged drug-induced gene expression signatures to engineer dynamic drug features by using a gating mechanism. Recently, a multi-type DDI prediction model named MDDI-SCL \cite{lin2022mddi} is presented by supervised contrastive learning and three-level loss functions.}

\subsubsection{Molecular graph}
Recent advances in artificial intelligence and technologies provide a set of potentially promising GNNs based approaches for drug-related prediction tasks, including molecular property \cite{wang2022molecular}, and molecular interactions \cite{li2022adaptive}. Naturally, drug molecules can be encoded by graph with atoms as nodes and chemical bonds as edges. Graph convolution neural networks (GCNs) have been proposed to extract node-level or graph-level features in various constructed graph \cite{duvenaud2015convolutional}. For example, MR-GNN \cite{ijcai2019p551} is proposed to use multiple graph convolution layers to extract node features from different neighboring nodes in a structured entity graph. Moreover, MHCADDI \cite{deac2019drug} leverages a co-attentional mechanism to combine the type of side-effect and the molecular structures to obtain drug-level representation. EPGCN-DS \cite{sun2020structure} adopts a GCN based framework for type-specific DDI identification from molecular structures. GNN-DDI \cite{feng2022prediction} learns k-hops drug representations its molecular graph via a five-layer GAT encoder. MFFGNN \cite{he2022multi} combines the topological structure in molecular graphs with the interaction relationship between drugs and the local chemical context in SMILES sequences. Furthermore, Molormer \cite{zhang2022molormer} takes the two-dimension (2D) structures of drugs as input and encodes the molecular graph with spatial information based on a lightweight attention mechanism. DeepDrug \cite{yin2022deepdrug} captures the intrinsic structural information of a compound by utilizing relational GCN module. Recently, R$^2$-DDI \cite{lin2022r2} further learns the drug representation by designing a relation-aware feature refinement framework.

\subsubsection{Substructure based}
\textcolor{black}{Different from the aforementioned methods (i.e., MR-GNN) that takes the whole chemical structures into account, more recent efforts have attempted to leverage GNN for powerful feature extraction of drug substructures. A chemical substructure representation framework named CASTER \cite{huang2020caster} encodes the functional substructures of drugs. SSI-DDI \cite{nyamabo2021ssi} operates directly on the raw molecular graph representations to identify pairwise interactions between their corresponding substructures. A gated MPNN (GMPNN) \cite{nyamabo2022drug} learns chemical substructures with different sizes and shapes from the molecular graph representations. A substructure-aware tensor model, referred as to STNN-DDI \cite{yu2022stnn}, learns a 3D tensor to characterize a substructure-substructure interaction space. SA-DDI \cite{yang2022learning} develops a directed MPNN with attention mechanism to extract the size- and shape-adaptive substructures. A Transformer-like framework (MSAN) \cite{zhu2022molecular} extracts substructures via attention mechanism to associate atoms with learnable pattern vectors. DSN-DDI \cite{li2023dsn} employs local and global representation learning modules iteratively, and learns drug substructures from intra-view and inter-view simultaneously. DGNN-DDI \cite{ma2023dual} exploits the molecular structure and interaction information between chemical substructure via a co-attention mechanism. Furthermore, incorporating geometric information into GNNs to benefit some molecular prediction tasks has recently gained research attention \cite{fang2022geometry}, and 3D structures of drug molecules also contribute to DDIs tasks, where 3DGT-DDI \cite{he20223dgt} adopts 3D structural information of molecular graph and position information to improve the model performance, which can deeply explore the effect of drug substructure on DDI relationship.}

\subsection{\textit{\textbf{Network based methods}}}
\textcolor{black}{In general, network based approaches infer the novel DDIs via label propagation \cite{zhang2015label, park2015predicting}, multiple sources \cite{cami2013pharmacointeraction} or newly calculated features \cite{zhang2019sflln}. With the increasing availability of large biomedical network and the rapid development of deep learning, some studies attempt to incorporate with various advanced techniques, including graph embedding, link prediction, and knowledge graph.}
\subsubsection{Graph embedding}
Various graph embedding algorithms have been proposed to acquire potentially effective network-based features, including matrix factorization-based methods (e.g., GraRep \cite{cao2015grarep}, HOPE \cite{ou2016asymmetric}, FGRMF \cite{zhang2018feature}, BRSNMF \cite{shi2019detecting}) that utilize the adjacency matrix as the input to learn latent embeddings from matrix factorization and random walk-based methods (e.g., DeepWalk \cite{deepwalk14}, node2vec \cite{grover2016node2vec} and struc2vec \cite{ribeiro2017struc2vec}) that first generate sequences of nodes through random walks and then feed the sequences into the model to learn node representations, and neural network-based methods (e.g., Line \cite{tang2015line}, SDNE \cite{wang2016structural} and GAE \cite{kipf2016variational}) that adopt different neural architectures and use various graph information as input. More detailed introduction of graph embedding on biomedical networks refer to \cite{yue2020graph}. Recently, GCNMK \cite{wang2022predicting} obtains the embeddings of drugs by constructing two DDI graphs as the graph kernels.

\subsubsection{Link prediction}
Meanwhile, some GNN approaches cast the prediction as a link prediction problem on DDI graph or network. Previously, Decagon \cite{zitnik2018modeling} is presented to develop a graph auto-encoder approach for multirelational link prediction on a multi-modal graph that consists of multiple interactions (e.g., drug-protein target interactions). In addition to aggregate information from direct interactions in biological network, a skip similarity approach named SkipGNN \cite{huang2020skipgnn} that receives neural messages from two-hop neighbors and direct neighbors in the interaction network. Analogously, DANN-DDI \cite{liu2022enhancing} builds multiple drug feature networks and learns drug representations from these networks by using the graph embedding method. HOGCN \cite{kishan2021predicting} is introduced to adopt a higher-order GCN to gather different features from the higher-order neighborhood for biomedical interaction prediction. A GNN based on graph structure and initial features, named LR-GNN \cite{kang2022lr}, constructs the link representation by designing a propagation algorithm to capture the node embedding. Furthermore, DGAT-DDI \cite{feng2022directed} is the first approach for predicting asymmetric interactions among drugs, and it designs a directed GAT to learn the embeddings of the source and the target role. deepMDDI \cite{feng2022deepmddi} learns the topological features of DDI network by combining RGCN encoder with similarity regularization of multiple drug features. Recently, a relation-aware network embedding model, abbreviated RANEDDI \cite{yu2022raneddi}, extracts the multirelational information and relation-aware network structure information together.

\subsubsection{Knowledge graph}
Existing GNN approaches for DDIs typically depend on one source of information, while using information from multiple sources could help improve predictions \cite{hu2021hiscf, su2022attention}. Particularly, knowledge graph (KG) has greatly stimulated research on various domains, including relation inference and recommendation \cite{wang2019knowledge}. In our knowledge, KG-DDI \cite{karim2019drug} is the first specialized for DDIs task that embeds the nodes in the constructed KG using various embedding approaches. Another KG embedding framework (AAEs) \cite{dai2021drug} uses adversarial autoencoders based on Wasserstein distances and Gumbel-Softmax relaxation. Furthermore, KGNN \cite{lin2020kgnn} successfully adopts GCNs with neighborhood sampling to explicitly extract the neighborhood relations. More recently, subgraph structures have been found to contain rich information for many graph learning tasks. SumGNN \cite{yu2021sumgnn} further uses KG to extract tractable pathway by designing a graph summarization module on subgraphs. And a link-aware graph attention method called LaGAT \cite{hong2022lagat} generates multiple attention pathways for drug entities based on various drug pair links in KG. DDKG \cite{su2022attention} further learns the drug embeddings from their attributes in the KG, and then simultaneously considers both neighboring node embeddings and triple facts by attention mechanism. KG2ECapsule \cite{su2022biomedical} integrates capsule network to explicitly model the multi-relational DDI data based on biomedical KG.

\subsection{\textit{\textbf{NLP based methods}}}
As training DNNs from scratch often requires a large number of labeled data which are expensive to acquire in real-world scenarios, inspired by the recent success in \textcolor{black}{natural language processing (NLP)}, pre-trained models have been proposed to learn universal molecular representations from massive unlabeled molecules and fine-tuned on downstream tasks with task-specific labeled data. BioBERT \cite{lee2020biobert} \textcolor{black}{is} first introduced to investigate how the pre-trained language model BERT \cite{devlin2018bert} are pre-trained using large-scale unlabeled molecular databases and then fine-tuned \textcolor{black}{for adaption to} biomedical text mining. \textcolor{black}{Subsequently} tremendous efforts have been devoted to pre-trained language model for biomedical prediction tasks, including property prediction \cite{fang2022geometry}, molecular generation \cite{bagal2021molgpt}, peptide and HLA (pHLA) binding prediction \cite{chu2022transformer}. More systematic introduction of molecular pre-trained models refer to \cite{xia2022systematic}.

\subsection{\textit{\textbf{Hybrid methods}}}
\textcolor{black}{Despite the remarkable progress gained by previous methods, improving the prediction accuracy is still crucial. Hybrid methods is proposed to combine with two or multiple types of existing methods in an efficient pattern.} For example, GoGNN \cite{ijcai2020p183} extracts features from both structured entity graphs and DDI network in a hierarchical way via dual-attention mechanism. MUFFIN \cite{chen2021muffin} jointly learns the drug representation from molecular structure and biomedical KG. BioDKG-DDI \cite{ren2022biodkg} adaptively integrates three different types of drug features, molecular structure, drug global information and drug functional similarity representation to predict novel DDIs. Recently, contrastive learning has been successfully applied in the application of bioinformatics, including gene regulatory interactions \cite{zheng2022accurate}, and drug-target interaction \cite{li2022supervised}. A novel unsupervised contrastive learning method named MIRACLE \cite{wang2021multi} is introduced, and it treats a DDI network as a multi-view graph where each node in the interaction graph represents a drug molecular graph instance. AMDE \cite{pang2022amde} jointly encodes 2D graph feature and 1D SMILES sequence by using message passing attention network and Transformer, respectively.

As illustrated in Table \ref{tab1} in chronological order, we present different deep and graph learning methods. Specifically, the columns of \textit{Model}, \textit{Input}, \textit{Representation}, \textit{Architecture}, \textit{Task}, and \textit{Code} represent the name, data format as the input of model, encoding form of input data, detailed architecture or technology adopted by the proposed model, functions can be implemented by the model, and available link of source code, respectively. The methods listed in Table \ref{tab1} are also appeared in the taxonomy in Figure \ref{fig1}.

\section{Discussion}
To comprehensively investigate the predictive performance of deep and graph learning models, we compared the experimental results of surveyed methods under binary, multi-class and multi-label classification tasks, respectively. In the following sections, we first introduce the benchmark datasets, then present the detailed of evaluation metrics under different prediction tasks, and finally analyze the comparison results.

\subsection{\textit{\textbf{Benchmark dataset}}}
We chose DrugBank and TWOSIDES datasets as benchmark datasets owing to their wide use across many studies. For the binary classification task, it always assigned a label ''1'' or ''0'' to indicate whether an interaction occurs between each pair of drugs in DrugBank and TWOSIDES datasets. For the multi-class classification task, the DrugBank dataset contains 191,808 DDI triplets with 1,706 drugs and 86 types of pharmacological relationships between drugs \cite{li2023dsn}. Following the same criterion in \textit{Decagon} \cite{zitnik2018modeling}, the interaction types with $<$500 triplets were removed, resulting in 4,576,287 DDI triplets with 963 interaction types in the TWOSIDES dataset. For the multi-label classification task, the TWOSIDES dataset contains 645 drugs (nodes) and 46,221 drug-drug pairs (edges) with 200 different drug side effect types as labels. For each edge, it may be associated with multiple labels. Following \textit{Decagon} \cite{zitnik2018modeling}, it kept 200 commonly-occurring DDI types ranging from Top-600 to Top-800 to ensure every DDI type has at least 900 drug combinations. As reported in \textit{KG2ECapsule} \cite{su2022biomedical}, it extracted the drug relation from the \textit{description} of \textit{drug-interaction} in DrugBank dataset. The aim was to categorize the types of DDI relations into two groups based on this extracted information. Following the same data split scheme in GMPNN \cite{nyamabo2022drug}, the benchmark dataset was split into train, validation, and test sets using a ratio of 6:2:2. Negative samples were randomly generated at a ratio of 1:1, meaning that they consisted of drug pairs that had not appeared in the positive samples.

\begin{table*}[htpb]
\caption{Performance evaluation under binary classification task.} \label{tab3}%
\centering
\begin{tabular}{c c c c c c c c}
\toprule
Method & Year & AUPRC & ACC & AUROC & F1 & Remarks \\
\midrule
\textbf{Dataset 1: DrugBank$^a$} \\
DeepWalk \cite{deepwalk14}      & 2014 & 0.9070 & 0.8349 & 0.9181 & 0.8357 & Network based method\\
GreRep \cite{cao2015grarep}     & 2015 & 0.9115 & 0.8443 & 0.9230 & 0.8461 & Network based method\\
LINE \cite{tang2015line}        & 2015 & 0.8915 & 0.8280 & 0.9092 & 0.8318 & Network based method\\
SDNE \cite{wang2016structural}  & 2016 & 0.8782 & 0.8303 & 0.9029 & 0.8373 & Network based method\\
GAE \cite{kipf2016variational}  & 2016 & 0.7403 & 0.7491 & 0.8085 & 0.7889 & Network based method\\
struc2vec \cite{ribeiro2017struc2vec}  & 2017   & 0.8672 & 0.7882 & 0.8735 & 0.7962 & Network based method\\
KG-DDI \cite{karim2019drug}     & 2019 &  -     & 0.7867 & 0.7867 & 0.7843 & Network based method\\
KGNN \cite{lin2020kgnn}         & 2020 & 0.9892 & 0.9561 & 0.9912 & 0.9566 & Network based method\\
AAEs \cite{dai2021drug}         & 2021 & 0.7899 & -      & 0.9480 & -      & Network based method\\
DANN-DDI \cite{liu2022enhancing}& 2022 & 0.9709 & \textbf{0.9962} & 0.9763 & 0.9692 & Network based method\\
DGAT-DDI \cite{feng2022directed}& 2022 & 0.943  & 0.886  & 0.951  & 0.884  & Network based method\\
RANEDDI \cite{yu2022raneddi}    & 2022 & \textbf{0.9894} & -      & 0.9898 & 0.9562 & Network based method\\
AMDE \cite{pang2022amde}        & 2022 & -      & 0.9763 & 0.9901 & 0.9760 & Network based method\\
DeepDDI \cite{ryu2018deep}      & 2018 & 0.828  & -      & 0.844  & 0.772  & Chemical structure based method\\
KMR \cite{shen2019kmr}          & 2019 & 0.9568 & 0.9219 & 0.9512 & 0.9191 & Chemical structure based method\\
CASTER \cite{huang2020caster}   & 2020 & 0.829  & -      & 0.861  & 0.796  & Chemical structure based method\\
SSI-DDI \cite{nyamabo2021ssi}   & 2021 & 0.9814 & 0.9447 & 0.9838 & -      & Chemical structure based method\\
MFFGNN \cite{he2022multi}       & 2022 & 0.9681 & -      & 0.9539 & 0.9254 & Chemical structure based method\\
DeepDrug \cite{yin2022deepdrug} & 2022 & 0.98   & -      & -      & 0.94   & Chemical structure based method \\
GMPNN \cite{nyamabo2022drug}    & 2022 & -      & 0.9530 & 0.9846 & -      & Chemical structure based method\\
SA-DDI \cite{yang2022learning}  & 2022 & -      & 0.9623 & 0.9880 & 0.9629 & Chemical structure based method\\
MSAN \cite{zhu2022molecular}    & 2022 & -      & 0.9700 & 0.9927 & 0.9704 & Chemical structure based method\\
R$^2$-DDI \cite{lin2022r2}      & 2022 & -      & 0.9815 & \textbf{0.9970} & \textbf{0.9816} & Chemical structure based method\\
3DGT-DDI \cite{he20223dgt}      & 2022 & -      & -      & 0.970  & -      & Chemical structure based method\\
DSN-DDI \cite{li2023dsn}        & 2023 & -      & 0.9694 & 0.9947 & 0.9693 & Chemical structure based method\\
MIRACLE \cite{wang2021multi}    & 2021 & 0.9234 & -      & 0.9551 & 0.8360 & Hybrid method \\
BioDKG-DDI \cite{ren2022biodkg} & 2022 & -      & 0.9370 & 0.9830 & 0.9390 & Hybrid method\\
\textbf{Dataset 2: TWOSIDES$^b$} \\
MR-GNN \cite{ijcai2019p551}     & 2019 & -      & 0.7623 & 0.85   & 0.7788 & Chemical structure based method \\
MHCADDI \cite{deac2019drug}     & 2019 & -      & -      & 0.8820 & -      & Chemical structure based method \\
SSI-DDI \cite{nyamabo2021ssi}   & 2021 & -      & 0.7820 & 0.8585 & 0.7981 & Chemical structure based method \\
DeepDrug \cite{yin2022deepdrug} & 2021 & -      & -      & -      & 0.84   & Chemical structure based method \\
GMPNN \cite{nyamabo2022drug}    & 2022 & -      & 0.8283 & 0.9007 & 0.8408 & Chemical structure based method \\
SA-DDI \cite{yang2022learning}  & 2022 & -      & 0.8745 & 0.9317 & 0.8835 & Chemical structure based method\\
R$^2$-DDI \cite{lin2022r2}      & 2022 & -      & 0.8615 & 0.9149 & 0.8731 & Chemical structure based method\\
DSN-DDI \cite{li2023dsn}        & 2023 & -      & \textbf{0.9883} & \textbf{0.9990} & \textbf{0.9883} & Chemical structure based method\\
\botrule
\end{tabular}
\begin{tablenotes}
$^a$ The performance on DrugBank dataset of DeepDDI was directed from CASTER results, and that of DeepWalk, GreRep, LINE, SDNE, GAE, struc2vec and KG-DDI were reported from KGNN results, and that of other methods were directly obtained from original papers. The division of the train and test set might be different for each model.\\
$^b$ The performance on TWOSIDES dataset of MR-GNN, MHCADDI, SSI-DDI, GMPNN and SA-DDI were reported from DSN-DDI results, and that of other methods were directly obtained from original papers. The division of the train and test set might be different for each model.
\end{tablenotes}
\end{table*}

\subsection{\textit{\textbf{Evaluation metrics}}}
Generally, we denote the true label and predicted values of DDIs by $y$ and $\hat{y}$, respectively. For the binary classification prediction, experiment results are reported with the following four metrics across the 5-folds. \textit{Area under the precision-recall curve (AUPRC)} is the area under the plot of the precision rate against recall rate at various thresholds, \textit{accuracy (ACC)} is defined as the number of correct predictions divided by the number of total predictions, \textit{area under the receiver operating characteristic (AUROC)} is the area under the plot of the true positive rate against the false positive rate at various thresholds, and \textit{F1 score} is the harmonic mean of precision and recall. The corresponding mathematical calculation is represented as follows.

\begin{equation}
    Precision = \frac{TP}{TP+FP},
\end{equation}
\begin{equation}
    Recall = TPR =  \frac{TP}{TP+FN},
\end{equation}
\begin{equation}
FPR = \frac{FPR}{FP+TN}
\end{equation}
\begin{equation}
    ACC = \frac{TP+TN}{TN+TP+FN+FP},
\end{equation}
\begin{equation}
    F1-score = \frac{2TP}{2TP+FN+FP},
\end{equation}
where $TP$, $FP$, $TN$ and $FN$ denote the value of true positive, false positive, true negative and false negative, respectively. The $AUPRC$ curve is drawn based on the values of $FPR$ and $TPR$, where the x-axis is $TPR$ and the y-axis is $FPR$. This is in contrast to $AUROC$ curves, where the x-axis is $FPR$ and the y-axis is $TPR$.

For the multi-class classification prediction, we follow DeepDDI \cite{ryu2018deep} and consider the following metrics, including \textit{mean accuracy}, \textit{macro precision}, \textit{macro recall} and \textit{macro F1}. Macro metrics are used to reflect the average performance
across different interaction types. For example, \textit{macro precision} is defined as the average of the precision values of different interaction
types. Their definitions are as follows:

\begin{equation}
    Mean\ accuracy = \frac{1}{l}\sum_{i=1}^{l}\frac{TP_i+TN_i}{TP_i+FN_i+FP_i+TN_i},
\end{equation}
\begin{equation}
    Macro\ recall = \frac{1}{l}\sum_{i=1}^{l}\frac{TP_i}{TP_i+FN_i},
\end{equation}
\begin{equation}
    Macro\ precision = \frac{1}{l}\sum_{i=1}^{l}\frac{TP_i}{TP_i+FP_i},
\end{equation}
\begin{equation}
    Macro\ F1 = \frac{2(Macro\ precision)(Macro\ recall)}{(Macro\ precison) + (Macro\ recall)},
\end{equation}
where $l$ is the number of DDI interaction types. In addition, to considering both \textit{precision} and \textit{recall}, we selected the threshold value, which achieves the maximum value of \textit{F1} in each interaction type, as the type-specific threshold.

For multi-label classification prediction, we follow \textit{SumGNN} \cite{yu2021sumgnn} and a group of metrics is used to measure the prediction, including \textit{ROC-AUC}, \textit{PR-AUC}, \textit{Accuracy (ACC)} and \textit{F1-score}. \textit{ROC-AUC} is the average area under the receiver operating characteristic curve as $ROC-AUC = \sum_{k=1}^{n}TP_k \Delta FP_k$, where $k$ represents $k$th true-positive and false-positive operating point ($TP_k$, $FP_k$). \textit{PR-AUC} is the average area under precision-recall curve $PR-AUC= \sum_{k=1}{n} \Delta Rec_k$, where $k$ is $k$th precision/recall oprating point ($Prec_k$, $Rec_k$). For each side effect type, the performance is individually calculated and use the average performance over all side effects as the final result.

\begin{table*}[htpb]
\renewcommand{\thetable}{4}
\caption{Performance evaluation under multi-class classification task.} \label{tab4}%
\centering
\begin{tabular}{c c c c c c c c}
\toprule
Method & Year & Mean accuracy & Macro precision & Macro recall & Macro F1 & Remarks \\
\midrule
\textbf{Dataset 1: DrugBank$^a$} \\
DeepWalk \cite{deepwalk14}        & 2014  & 0.8000  & 0.8220  & 0.7101  & 0.7469  & Network based method\\
LINE \cite{tang2015line}          & 2015  & 0.7506  & 0.6870  & 0.5451  & 0.5804  & Network based method\\
Decagon \cite{zitnik2018modeling} & 2018  & 0.8719  & -       & -       & 0.5735  & Network based method\\
KG-DDI \cite{karim2019drug}       & 2019  & 0.8923  & 0.7945  & 0.7667  & 0.7666  & Network based method\\
KGNN \cite{lin2020kgnn}           & 2020  & 0.9127  & 0.8583  & 0.8170  & 0.8291  & Network based method\\
SkipGNN \cite{huang2020skipgnn}   & 2020  & 0.8583  & -       & -       & 0.5966  & Network based method\\
SumGNN \cite{yu2021sumgnn}        & 2021  & 0.9266  & -       & -       & 0.8685  & Network based method\\
LaGAT \cite{hong2022lagat}        & 2022  & 0.9604  & -       & -       & 0.9289  & Network based method\\
DeepDDI \cite{ryu2018deep}        & 2018  & 0.8371  & 0.7275  & 0.6611  & 0.6848  & Chemical structure based method\\
DDIMDL \cite{deng2020multimodal}  & 2020  & 0.8852  & 0.8471  & 0.7182  & 0.7585  & Chemical structure based method\\
SSI-DDI \cite{nyamabo2021ssi}     & 2021  & 0.8965  & 0.8763  & 0.9321  & 0.8993  & Chemical structure based method\\
GMPNN \cite{nyamabo2022drug}      & 2022  & 0.9485  & 0.9346  & 0.9725  & 0.9495  & Chemical structure based method\\
SA-DDI \cite{yang2022learning}    & 2022  & 0.9565  & 0.9472  & 0.9746  & 0.9573  & Chemical structure based method\\
Molormer \cite{zhang2022molormer} & 2022  & \textbf{0.9667}  & 0.9419  & 0.9270  & 0.9311  & Chemical structure based method\\
MDDI-SCL \cite{lin2022mddi}       & 2022  & 0.9378  & 0.8804  & 0.8767  & 0.8755  & Chemical structure based method\\
DGNN-DDI \cite{ma2023dual}        & 2023  & 0.9609  & 0.9472  & \textbf{0.9788}  & \textbf{0.9616}  & Chemical structure based method\\
MUFFIN \cite{chen2021muffin}      & 2021  & -       & \textbf{0.9648}  & 0.9495  &         & Hybrid method method\\
\botrule
\end{tabular}
\begin{tablenotes}
$^a$ The performance on DrugBank dataset of DeepWalk, LINE, DeepDDI, KG-DDI and KGNN were reported from MUFFIN results, and that of Decagon and SkipGNN were obtained from SumGNN results, and that of GMPNN, SA-DDI and SSI-DDI were obtained from DGNN-DDI results, and that of other methods were directly obtained from original papers.
\end{tablenotes}
\end{table*}

\subsection{\textit{\textbf{Results}}}
In this section, we compared state-of-the-art deep and graph learning models under binary, multi-class and multi-label classification prediction task, respectively. Table \ref{tab3} shows the comparison results of 30 models under binary classification task on two benchmark datasets. The performance of DDIs prediction achieved by these models were all measured in terms of AUPR, ACC, AUROC and AUC under 5-fold cross-validation. The greater these evaluation metrics the better the prediction. Although the division of the training and test sets could be specific to models, such an evaluation is still statistically significant. Specifically, from the observation we found that RANEDDI (AUPRC = 0.9894) and KGNN (AUPRC = 0.9892), which belong to network based methods, achieve the best and second-best AUPRC performance compared with chemical structure based and hybrid methods on DrugBank datasets. This is because these methods (i.e., RANEDDI and KGNN) can explore multi-relational information contained in the DDI network or knowledge graph, while the graph embedding approaches like DeepWalk, GraRep, DeepDDI or substructure based method (e.g., CASTER) only learn from similar drug features or chemical structural information. Note that DANN-DDI obtains the best ACC result of 0.9962 over all models, and R$^2$-DDI achieves the best performance in terms of AUROC and F1 score. Meanwhile, experimental results on TWOSIDES dataset show that DSN-DDI, recently published chemical structure based model, achieves better performance than other baseline models on all evaluation metrics. Particularly, the ACC, AUROC and F1 result of DSN-DDI is 0.9883, 0.9990 and 0.9883, respectively. Interestingly, the comparison indicates that network based methods (e.g., DANN-DDI and RANE-DDI) show similar performance to chemical structure based methods (e.g., R$^2$-DDI), while hybrid methods show stable performance on DrugBank datasets under binary classification task.

In the multi-class classification task, we chose the DrugBank dataset as benchmark dataset owing to its wide use across many studies and collected 10 deep and graph learning models into the comparison list, which is shown in Table \ref{tab4}. From this table, we found that chemical structure based methods significantly outperform network based and hybrid methods on most metrics. More specifically, Molormer achieved the best score of 0.9667 on mean accuracy, DGNN-DDI achieved the macro recall and macro F1 score of 0.9788 and 0.9616 compared to other methods, respectively. In addition, we can see that chemical structure based methods achieved stable performances across all metrics. For example, the macro recall of GMPNN and SA-DDI are 0.9725 and 0.9746, respectively. These results demonstrate that they achieved similar performance with DGNN-DDI, indicating that they belong to substructure based methods. This is a very encouraging result. The reason could be that (i) DDIs are fundamentally caused by chemical substructure interactions, especially in multi-class classification tasks that focus on atom similarity and key substructures; (ii) more effective strategies are proposed by these methods to specifically detect substructures with irregular size and shape, which can further enhance the representation capability of the model. In addition, the multi-class classification task is more difficult than the binary classification task.

In the multi-label classification task, we chose DrugBank and TWOSIDES datasets as benchmark datasets and compared 11 deep and graph learning models as shown in Table \ref{tab5}. From this table, we observed that KG2ECapsule and SumGNN consistently outperformed other methods in all evaluation metrics. In particular, KG2ECapsule improves over the strongest baselines with respect to PR-AUC by 2.71\%, ACC by 1.03\%, ROC-AUC by 2.8\% and F1 by 2\% on DrugBank dataset, respectively. The reason for this is that KG2ECapsule is capable of modeling the triplets and integrating the relations of edges into embedding. Meanwhile, on TWOSIDES dataset, SumGNN achieved at least 2.45\% on PR-AUC, 2.82\% on ROC-AUC higher performance than other methods. This justifies that SumGNN is more effective to harness the external knowledge via subgraphs. More interestingly, with comparison to other KG based methods (e.g., KGNN and KG-DDI), we found that KG2ECapsule and SumGNN can consistently outperform them on both datasets, which indicates that simply adopting KG embeddings as well as neighborhood sampling are insufficient to fully harness the KG information for DDIs prediction. Moreover, network based methods achieved better performances in the multi-label classification task.

\begin{table*}[htpb]
\renewcommand{\thetable}{5}
\caption{Performance evaluation under multi-label classification task.} \label{tab5}%
\centering
\begin{tabular}{c c c c c c c c}
\toprule
Method & Year & PR-AUC & ACC & ROC-AUC & F1 & Remarks \\
\midrule
\textbf{Dataset 1: DrugBank$^a$} \\
DeepWalk \cite{deepwalk14}          & 2014 & 0.4782 & 0.6163 & 0.6501 & 0.5861 & Network based method\\
LINE \cite{tang2015line}            & 2015 & 0.4923 & 0.6374 & 0.6926 & 0.6190 & Network based method\\
KGNN \cite{lin2020kgnn}             & 2020 & 0.8587 & 0.7947 & 0.8602 & 0.7945 & Network based method\\
KG2ECapsule \cite{su2022biomedical} & 2023 & \textbf{0.8858} & \textbf{0.8050} & \textbf{0.8882} & \textbf{0.8145} & Network based method\\
\textbf{Dataset 2: TWOSIDES$^b$} \\
DeepWalk \cite{deepwalk14}          & 2014 & 0.6160 & -      & 0.8708 & -      & Network based method\\
LINE \cite{tang2015line}            & 2015 & 0.6043 & -      & 0.8621 & -      & Network based method\\
node2vec \cite{grover2016node2vec}  & 2016 & 0.8887 & -      & 0.9066 & -      & Network based method\\
Decagon \cite{zitnik2018modeling}   & 2018 & 0.9060 & -      & 0.9172 & -      & Network based method\\
KG-DDI \cite{karim2019drug}         & 2019 & 0.6527 & -      & 0.8906 & -      & Network based method\\
KGNN \cite{lin2020kgnn}             & 2020 & 0.6584 & -      & 0.8948 & -      & Network based method\\
SkipGNN \cite{huang2020skipgnn}     & 2020 & 0.9090 & -      & 0.9204 & -      & Network based method\\
SumGNN \cite{yu2021sumgnn}          & 2021 & \textbf{0.9335} & -      & \textbf{0.9486} & -      & Network based method\\
DeepDDI \cite{ryu2018deep}          & 2018 & 0.5032 & -      & 0.8301 & -      & Chemical structure based method\\
MUFFIN \cite{chen2021muffin}        & 2021 & 0.7033 & -      & 0.9160 & -      & Hybrid method\\
\botrule
\end{tabular}
\begin{tablenotes}
$^a$ The performance on DrugBank dataset of DeepWalk, LINE and KGNN were reported from KG2ECapsule results.\\
$^b$ The performance on TWOSIDES dataset of DeepWalk, LINE, node2vec, Decagon, KG-DDI, KGNN and SkipGNN were reported from SumGNN results, and that of DeepDDI was obtained from MUFFIN results, and that of other methods were directly obtained from original papers.
\end{tablenotes}
\end{table*}

\section{Challenge and Opportunities}
Deep and graph learning techniques have distinct advantages over traditional machine learning methods in tackling the computational drug discovery. Although many studies focus on the prediction of DDIs and high prediction performance have been proposed, there still remains several challenges and promising future directions as follows.

\subsection{\textit{\textbf{Dataset imbalance}}}
Most deep and graph learning models in drug discovery pipeline need large amounts of data for model training and validation. The lack of enough known DDIs and experimentally validated negative samples are major obstacles for deep and graph learning models to have positive influence on DDIs prediction, especially in the application of real-world scenarios. For example, the imbalanced data for different relations in certain case are very sparse \textcolor{black}{with respect to} side-effect type, which will lead to poor generalizability for model performance. Meanwhile, current DDI benchmark datasets only include a small number of labeled (resp., positive) samples, in which the quality of data is not guaranteed and the dataset might be imbalanced for the lack of negative samples of drug-drug pair. As for unlabeled samples, most methods regard them as negative samples and sample the same number of negative drug pairs from non-interacting DDIs for model training. \textcolor{black}{These methods overlook} the fact that unlabeled samples may contain potential positive data, which would adversely influence the model performances. How to choose high-quality data and how to \textcolor{black}{address} insufficient training data remain challenges.

\subsection{\textit{\textbf{Multimodal representation}}}
The \textcolor{black}{potential of computational drug discovery lies in the variety} of multiple data modalities that provide complementary information \cite{luo2020multidimensional}. \textcolor{black}{Deep and graph} learning models using multimodal data will \textcolor{black}{have considerable} advantages over unimodal counterparts since the multimodal data offer complementary perspectives. Existing studies usually focus on the single modal data. For example, graph- or substructure-based method pay more attention to the molecular data containing structural information, while network-based approaches only consider the relationship between drug and relation in the drug level, neglecting the atom level of the pair interaction between drugs. These methods do not fully use other data modalities, such as drug-target interactions, drug-disease associations, protein pathways and evidences from electronic medical records, such information may be also highly related to DDIs and their induced adverse reactions. Thus, how to effectively \textcolor{black}{utilize} diverse and heterogeneous biological data is worth of \textcolor{black}{exploring}.

\subsection{\textit{\textbf{High-order drug associations}}}
Identifying the potential associations between drugs and related entities (e.g., diseases and microbe) is pivotal to understanding the underlying
disease mechanisms and facilitating personalized treatments. In the past few years, most methods have been proposed to concentrate on predicting pair-wise associations, such as drug-drug, drug-protein, drug-microbe and drug-disease interactions, these methods deal with them separately and fail to provide in-depth insights into high-order association patterns. For example, many diseases are closely related to various microbes, which interact with a variety of drugs in complex way, and the causal links between drugs, gut microbes and diseases require a workflow to uncover their intricate interactions. Such a workflow of triple-wise drug-microbe-disease associations can be regarded as high-order drug associations prediction. Meanwhile, high-order associations prediction is a fundamental task in multiple domains, including knowledge graphs, recommendation systems and bioinformatics. There is an urgent need to seek ways to develop effective methods for predicting high-order drug associations to speed up the process of drug discovery.

\subsection{\textit{\textbf{Model interpretability}}}
Deep and graph learning techniques offer great potential in many fields, but they are often essentially "black boxes" that are unable to provide confidence and actionability for the predicted results. As an essential process in drug discovery, DDIs prediction aims to identify and quantify the risks related to the usage of drugs for a better understanding of adverse drug effects and the pathogenic mechanisms. The latent embedding obtained by current \textcolor{black}{deep and graph} learning models is limited to capturing implicit correlations of the data, which \textcolor{black}{is} hard to provide reasonable explanations for the predicted interactions. Thus, the idea model \textcolor{black}{should} understand how the algorithms are constructed, what each layer learns, and what the embeddings represent. Meanwhile, interpretability and evidence support are \textcolor{black}{essential} for prediction methods in biomedical applications. It is also worthwhile to further focus on interpretability and to improve the reliability of predicted results.

\subsection{\textit{\textbf{Generative AI Models}}}
Recent advances in generative AI models, such as ChatGPT\footnote{\url{https://openai.com/blog/chatgpt}}, have shown remarkable success on a variety of domains. From Transformer to BERT to ChatGPT, the continuous advancement of generative AI models has opened up a new era of AI. These generative AI models are trained on large-scale datasets, providing a reasonable parameter initialization for a wide range of downstream applications, including natural language processing \cite{brown2020language}, computer vision \cite{dosovitskiy2020image}, and graph learning \cite{yun2019graph}. Moreover, generative AI models have been deployed in various stages of the drug development pipeline \cite{zeng2022deep}, ranging from AI-assisted target selection and validation to molecular design and chemical synthesis. In the near future, it is anticipated that generative AI models would be able to generate realistic data that can be used to identify potential DDIs. This data can then be utilized to improve existing models or create new models that are more effective at addressing the challenges mentioned above. By combining the power of generative AI models and advanced deep and graph learning techniques, it is conceivable to develop better models for predicting DDIs.

\begin{table*}[!t]
\caption{List of widely-used platform and toolkit for biomedical application.\label{tab4}}%
\centering
\begin{tabular}{cccccc}
\toprule
\textbf{Name}   & \textbf{Year}  & \textbf{Release}  & \textbf{Updates}  & \textbf{Application examples}    & \textbf{Website}\\
\midrule
DeepChem        & 2017           & V2.7.1            & $\surd$           &  \makecell[c]{Molecular property prediction\\Drug-target binding affinity prediction\\Physical properties prediction\\Protein structure analysis and descriptors extraction\\Number counting of cells in a microscopy image} & \href{https://deepchem.io}{\textcolor{blue}{Link}} \\ \hline
DeepPurpose     & 2020           & V0.1.5            & $\surd$           &  \makecell[c]{Drug target interaction prediction\\Drug property prediction\\Drug-drug interactions prediction\\Protein-protein interaction prediction\\Protein function prediction\\Antiviral drugs repurposing for SARS-CoV2 3CLPro\\Repurposing using customized training data} & \href{https://github.com/kexinhuang12345/DeepPurpose}{\textcolor{blue}{Link}} \\ \hline
PaddleHelix     & 2020           & V1.1.0            & $\surd$           &  \makecell[c]{Large-scale pre-training models of compounds and proteins\\Molecular property prediction\\Drug-target affinity prediction\\Molecular generation\\RNA design\\Drug-drug synergy prediction} & \href{https://github.com/PaddlePaddle/PaddleHelix}{\textcolor{blue}{Link}} \\ \hline
DGL-LifeSci     & 2021           & V0.3.1            & $\surd$           &  \makecell[c]{Property prediction\\Generative models\\Protein-ligand binding affinity prediction\\Reaction prediction} & \href{https://github.com/awslabs/dgl-lifesci}{\textcolor{blue}{Link}} \\ \hline
TorchDrug       & 2021           & V0.2.0            & $\surd$           & \makecell[c]{Property prediction\\ Pretrained molecular representations\\ Molecule generation\\ Retrosynthesis\\ Knowledge graph reasoning} & \href{https://torchdrug.ai}{\textcolor{blue}{Link}} \\ \hline
ADMETlab        & 2021           & V2.0              & $\surd$           & \makecell[c]{Absorption, Distribution, Metabolism, \\Excretion and Toxicity (ADMET) prediction} & \href{https://admetmesh.scbdd.com}{\textcolor{blue}{Link}} \\ \hline
ChemicalX       & 2022           & V0.1.0            & $\times$          & \makecell[c]{Drug-drug interactions prediction\\ Drug pair scoring task} & \href{https://github.com/AstraZeneca/chemicalx}{\textcolor{blue}{Link}} \\
\botrule
\end{tabular}
\end{table*}

\subsection{\textit{\textbf{Platform and toolkit}}}
To further \textcolor{black}{speed up the drug discovery process and enable more people with different scientific backgrounds to get involved in research}, many researcher and communities have been committed to the development of platform and toolkit based on machine learning and deep learning methods. Table \ref{tab4} illustrates the widely-used platform and toolkit for biomedical application. Specifically, DeepChem aims to provide a high quality open-source toolchain that \textcolor{black}{makes deep learning in drug discovery, materials science, quantum chemistry, and biology more accessible}. DGL-LifeSci is a DGL-based package for various \textcolor{black}{life science applications with graph neural networks and provides various functions}.
DeepPurpose is a deep learning-based molecular modeling and prediction toolkit involving many downstream tasks \textcolor{black}{, (e.g., compound property prediction, and protein function prediction)}. TorchDrug is a machine learning platform designed for drug discovery \textcolor{black}{that covers various techniques from GNNs, geometric deep learning, KGs, deep generative models, and} reinforcement learning. PaddleHelix is a bio-computing tool that takes advantage of the machine learning approaches, especially \textcolor{black}{DNNs}, for facilitating the development of the following areas, including drug discovery, vaccine design, and precision medicine. ADMETlab 2.0 \cite{xiong2021admetlab} is an \textcolor{black}{improved} version of the widely used ADMETlab\textcolor{black}{, which is used to} systematical evaluation of ADMET properties.
While fewer work are specialized for developing the platform or toolkit on DDIs prediction. To our knowledge, one such work named \textcolor{black}{ChemicalX} is a deep learning library for drug-drug interaction, polypharmacy side effect, and synergy prediction, \textcolor{black}{and also} includes state-of-the-art \textcolor{black}{DNN} architectures that solve the drug pair scoring task\textcolor{black}{, with implemented methods covering traditional SMILES sequence based techniques and MPNN} based models. \textcolor{black}{However, these} platforms and toolkits, which are mainly developed by individuals, do not have any maintenance or update schedule in place. \textcolor{black}{As a result, they will become increasingly obsolete} as the underlying programming framework and deep learning models continue to evolve.

\section{Conclusions and Outlook}
In this work, we provided a comprehensive review of deep and graph learning methods for drug-drug interactions prediction. We categorized existing approaches into traditional machine learning, deep learning and graph neural network (GNN)-based methods. We introduced data sources and summarized the widely-used molecular representation as well as some classic GNN model on DDIs prediction. To the end, we discussed the current challenges of existing deep and graph learning methods and suggested potential research directions for further development in DDIs prediction. In conclusion, the rapidly growth of deep and graph learning techniques has brought new opportunities for biomedical applications, including drug-related prediction tasks. However, the bottlenecks of these technologies, such as imbalance dataset, the issues of multimodal representation and high-order drug associations prediction, and the lack of or limited interpretability of the model impedes their application and further affects their prediction performance. Therefore, there is an urgent need to further develop and evaluate intelligent deep and graph learning models in realistic drug discovery scenarios in order to reach its full potential.

\section{Key Points}
\begin{itemize}
  \item \textcolor{black}{\textit{Structured taxonomy}. As shown in Fig. 1, we contribute a structured taxonomy to provide a broad overview of computational methods, which categorizes existing works from four perspectives: chemical structure based, network based, NLP based and hybrid methods.}
  \item \textcolor{black}{\textit{Current progress}. We systematically delineate the current research directions on the topic of deep and graph learning methods for DDIs prediction as illustrated in Table 1, and we further investigate the comparison performance of these representative baseline models as shown in Tables 3-5.}
  \item \textcolor{black}{\textit{Abundant resources.} We have gathered a comprehensive collection of resources dedicated to DDIs prediction. These collections include open-sourced deep and graph learning methods, available platform and toolkit, as well as an important paper list. These resources can be accessed our github\footnote{\url{https://github.com/xzenglab/resources-for-DDIs-prediction-using-DL}}, which will be continuously updated.}
  \item \textcolor{black}{\textit{Future directions}. We discuss the limitations of existing works and suggest several promising future directions.}
\end{itemize}

\section{Author Biographies}
\textbf{Xuan Lin} is currently a lecturer in the college of computer science, Xiangtan University (XTU), Xiangtan, China. His main research interests include machine learning, graph neural networks and bioinformatics.

\noindent\textbf{Lichang Dai} is an undergraduate in the college of computer science, Xiangtan University. His research interests are machine learning and bioinformatics.

\noindent\textbf{Yafang Zhou} is an undergraduate in the college of computer science, Xiangtan University. Her research interests are machine learning and data mining.

\noindent\textbf{Zu-Guo Yu} is a professor in Key Laboratory of Intelligent Computing and Information Processing of Ministry of Education, Xiangtan University. His research is focused on fractals, bioinformatics and complex networks and geomagnetic data analysis.

\noindent\textbf{Wen Zhang} is a professor in College of Informatics, Huazhong Agricultural University, China. His research interests include machine learning and bioinformatics.

\noindent\textbf{Jian-Yu Shi} is a Professor in the Northwestern Polytechnical University, Xi'an, China. His research interests include bioinformatics, cheminformatics and artificial intelligence.

\noindent\textbf{Dong-Sheng Cao} is currently a professor in the Xiangya School of Pharmaceutical Sciences, Central South University, China. His research interests can be found at the website of his group: http://www.scbdd.com.

\noindent\textbf{Li Zeng} is currently the head of AIDD department of Yuyao Biotech, Shanghai, China. He has published several research papers on Biochemical Pharmacology, The FASEB Journal, Scientific Reports, etc.

\noindent\textbf{Haowen Chen} is an associate professor with Hunan University. His research interests include bioinformatics and artificial intelligence. He has published several research papers in these fields including Npj Systems Biology, IEEE/ACM TCBB, Methods, COMPUTERS \& SECURITY etc.

\noindent\textbf{Bosheng Song} is an associate professor with the College of Information Science and Engineering, Hunan University, Changsha, China. His current research interests include membrane computing and bioinformatics.

\noindent\textbf{Philip S. Yu} (Fellow, IEEE) is a distinguished professor in computer science with the University of Illinois at Chicago and also holds the Wexler chair in information technology. His research interest is on Big Data, including data mining, data stream, database and privacy.

\noindent\textbf{Xiangxiang Zeng} (Senior Member, IEEE) is an Yuelu distinguished professor with the College of Information Science and Engineering, Hunan University, Changsha, China. His main research interests include computational intelligence, graph neural networks and bioinformatics.
\bibliographystyle{unsrt}
\bibliography{main}




\end{document}